\documentclass[lettersize,journal]{IEEEtran}
\usepackage{amsmath,amsfonts}
\usepackage{algorithmic}
\usepackage{algorithm}
\usepackage{array}
\usepackage[caption=false,font=footnotesize,labelfont=rm,textfont=rm]{subfig}
\usepackage{textcomp}
\usepackage{stfloats}
\usepackage{url}
\usepackage{verbatim}
\usepackage{graphicx}
\usepackage{cite}

\usepackage{textcomp}
\usepackage{multirow}
\usepackage{booktabs}
\usepackage{amssymb}
\usepackage{soul}
\usepackage{tablefootnote}
\usepackage{pifont}
\usepackage{threeparttable} 
\usepackage[table,xcdraw]{xcolor}
\usepackage{graphicx}
\usepackage{threeparttable}
\usepackage{tikz}

\definecolor{bl}{HTML}{0070c0}
\definecolor{rd}{HTML}{c00001}

\begin{document}

\title{Towards Faithful Class-level Self-explainability in Graph Neural Networks by Subgraph Dependencies
}

\author{Fanzhen Liu,
        Xiaoxiao Ma,
        Jian Yang,
        Alsharif Abuadbba,
        Kristen Moore,\\
        Surya Nepal,~\IEEEmembership{Fellow,~IEEE,}
        Cecile Paris,
        Quan Z. Sheng,
        and Jia Wu,~\IEEEmembership{Senior Member,~IEEE}
\thanks{F. Liu is with the School of Computing, Macquarie University, Sydney, NSW 2109, Australia, and also with CSIRO's Data61, Sydney, NSW 2015, Australia (e-mail: fanzhen.liu@mq.edu.au).}
\thanks{X. Ma, J. Yang, Q. Z. Sheng, and J. Wu are with the School of Computing, Macquarie University, Sydney, NSW 2109, Australia (e-mail: xiaoxiao.ma2@hdr.mq.edu.au; \{jian.yang, michael.sheng, jia.wu\}@mq.edu.au).}
\thanks{A. Abuadbba, K. Moore, S. Nepal, and C. Paris are with CSIRO's Data61, Sydney, NSW 2015, Australia (e-mail: \{sharif.abuadbba, kristen.moore, surya.nepal, cecile.paris\}@data61.csiro.au).}
}

\markboth{Journal of \LaTeX\ Class Files,~Vol.~x, No.~x, August~2025}%
{Shell \MakeLowercase{\textit{et al.}}: A Sample Article Using IEEEtran.cls for IEEE Journals}


\maketitle

\begin{abstract}
Enhancing the interpretability of graph neural networks (GNNs) is crucial to ensure their safe and fair deployment. Recent work has introduced self-explainable GNNs that generate explanations as part of training, improving both faithfulness and efficiency. Some of these models, such as ProtGNN and PGIB, learn class-specific prototypes, offering a potential pathway toward class-level explanations. However, their evaluations focus solely on instance-level explanations, leaving open the question of whether these prototypes meaningfully generalize across instances of the same class. In this paper, we introduce GraphOracle, a novel self-explainable GNN framework designed to generate and evaluate class-level explanations for GNNs. Our model jointly learns a GNN classifier and a set of structured, sparse subgraphs that are discriminative for each class. We propose a novel integrated training that captures graph–subgraph–prediction dependencies efficiently and faithfully, validated through a masking-based evaluation strategy. This strategy enables us to retroactively assess whether prior methods like ProtGNN and PGIB deliver effective class-level explanations. Our results show that they do not. In contrast, GraphOracle achieves superior fidelity, explainability, and scalability across a range of graph classification tasks. We further demonstrate that GraphOracle avoids the computational bottlenecks of previous methods—like Monte Carlo Tree Search—by using entropy-regularized subgraph selection and lightweight random walk extraction, enabling faster and more scalable training. These findings position GraphOracle as a practical and principled solution for faithful class-level self-explainability in GNNs.
\end{abstract}

\begin{IEEEkeywords}
Graph neural networks, class-level explainability, faithfulness, subgraph dependencies.
\end{IEEEkeywords}

\section{Introduction}
\label{sec:intro}
Graph neural networks (GNNs) have advanced the analysis of real-world data represented as {\it graphs}. As powerful tools for analyzing spatial structure and attribute distributions, GNNs have transformed graph data mining across a wide spectrum of applications. These range from fundamental tasks such as graph classification~\cite{xu2018how,yang2024graph}, link prediction~\cite{zhang2018link}, and community detection \cite{su2022survey,liu2020deep}, to specialized domains including molecular structure analysis and generation \cite{wang2022molecular, you2018graph}, drug discovery~\cite{jiang2021could}, social recommendation systems~\cite{fan2019gnn}, and fraud detection~\cite{liu2022eriskcom}.
However, their black-box nature poses a significant barrier to adoption in high-stakes domains. To enable transparency and trustworthy deployment, explaining how a GNN approaches its predictions is essential~\cite{kakkad2023survey}.

Recognizing the demand for human-interpretable models, researchers have developed a range of techniques to explore the relationship between GNN predictions and their underlying mechanisms \cite{yuan2023survey}.
Current research broadly falls into two main categories: 1) Instance-level explainers, which highlight input components (e.g., nodes, edges, or subgraphs) responsible for predictions on individual graph instances \cite{fang2023cooperative,wu2022discovering,ying2019GNNExplainer}; and 2) Class-level explainers, which aim to uncover patterns that are consistently predictive for an entire class of outputs \cite{azzolin2023global,yuan2020XGNN}.
While instance-level explanations provide useful localized insights, they are costly to verify at scale and rely heavily on expert (human) supervision. In contrast, class-level explanations distill higher-level patterns shared across instances, easing interpretability and reducing cognitive load for users\cite{yuan2023survey}. 

These class-level explanations are particularly valuable in domains where expert oversight is limited or the problem space is highly complex. By surfacing high-level, class-specific patterns, these models enhance both interpretability and practical utility. A common strategy for class-level explanation involves post-hoc explainer models, which analyze a trained GNN to construct prototypical graphs that characterize different classes \cite{yuan2020XGNN,wang2023gnninterpreter,wang2021towards}, or cluster instance-level explanations into shared patterns within classes \cite{azzolin2023global}.
However, these decoupled methods suffer from two key limitations: 1) they cannot leverage internal GNN representations learned during training, and 2) they risk capturing spurious or non-generalizable correlations rather than causal or semantically meaningful structures \cite{zhang2022protgnn}.

To address these limitations, recent research has explored {\it self-explainable} GNNs, which integrate explanation as part of the model directly into training. These models have primarily focused on instance-level explainability \cite{miao2022interpretable, wu2022discovering, liu2022graph, yang2025from}, enabling faithful instance-specific explanations while improving computational efficiency.
Among these, ProtGNN~\cite{zhang2022protgnn} and PGIB\cite{seo2023interpretable} introduce class-specific prototypes to connect the informative instance-specific subgraphs with class predictions. While these prototypes suggest potential for class-level insight, their evaluations remain exclusively instance-level, measuring how prototypes contribute to individual input predictions. Critically, neither method assesses whether its prototypes generalize across instances of the same class, nor whether they yield faithful or discriminative class-level explanations.

In this paper, we introduce GraphOracle, a novel framework designed for class-level self-explainabile GNNs. GraphOracle jointly learns a GNN classifier and a set of class-discriminative subgraphs that are sparse, representative, and faithful to the class predictions it explains. To enable meaningful and fair comparison with prior work, we propose a novel masking-based evaluation strategy that quantifies the faithfulness of class-level explanations. Using this, we retroactively evaluate ProtGNN and PGIB for class-level explanation efficacy and show they fall short-despite learning class-specific prototypes, they fail to produce generalizable or faithful class-level explanations.

\begin{figure}
    \centering
    \includegraphics[width=1\linewidth]{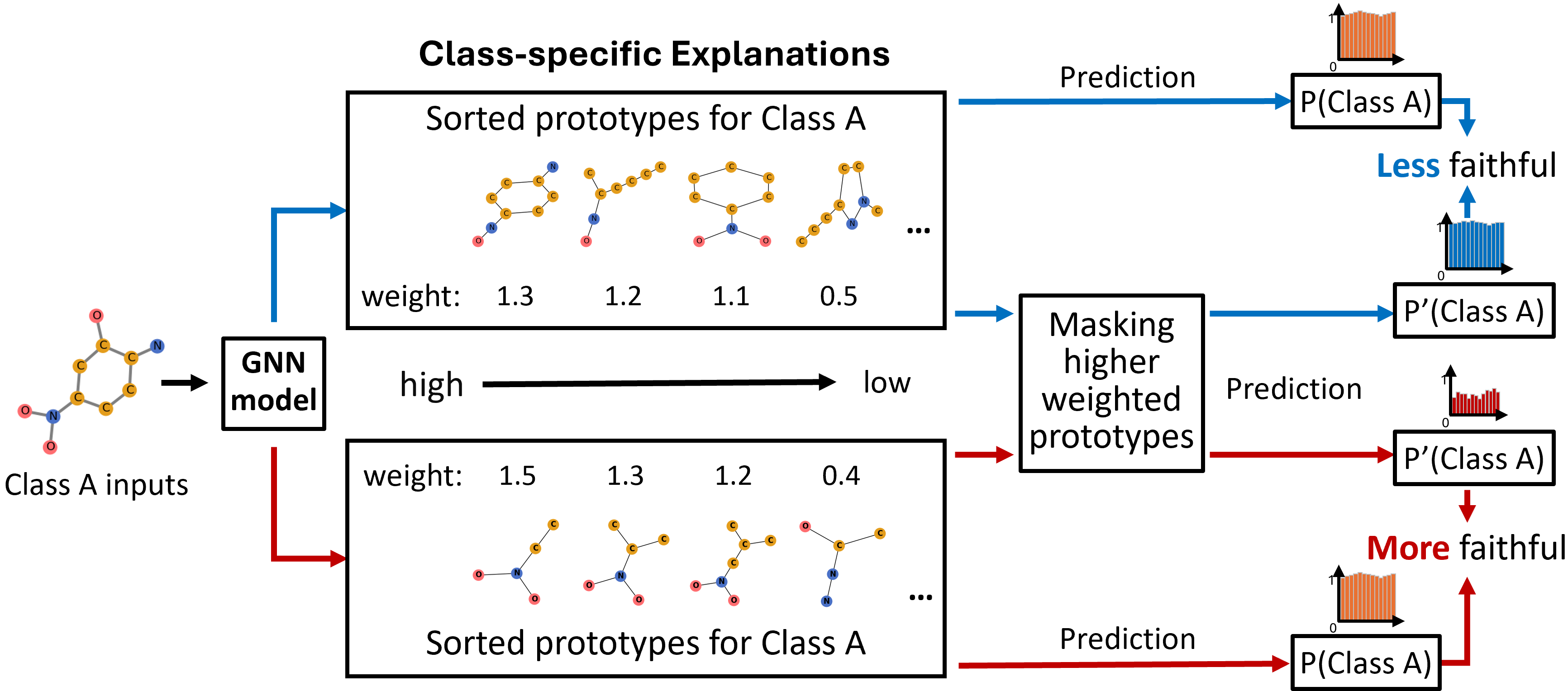}
    \caption{Assessing faithfulness of class-specific explanations. Faithful explanations should impact GNN predictions when key features are masked. The \textcolor{bl}{blue} GNN's predictions remain very little after masking highly-weighted prototypes identified by the explainer, indicating low faithfulness of the explanations. In contrast, the \textcolor{rd}{red} GNN's predictions change dramatically, suggesting more faithful explanations.}
    \label{fig:challenge1}
\end{figure}

One key issue for the existing methods is their inability to produce faithful class-specific explanations—i.e., critical subgraphs or prototypes that, when removed, cause significant changes in prediction (see Fig.~\ref{fig:challenge1} and Sec.~\ref{sec:effectiveness}).
In contrast, GraphOracle consistently produces more faithful and informative explanations across multiple datasets.
In addition to its improved faithfulness, GraphOracle addresses a second major challenge of existing self-explainable methods: efficiency. Prototype-based models like ProtGNN and PGIB rely on expensive Monte Carlo Tree Search to discover subgraph matches during training, incurring high computational overhead (see Sec.~\ref{sec:efficiency}). In contrast, GraphOracle introduces a novel training scheme that enables efficient, end-to-end learning of graph–subgraph–prediction dependencies using pre-extracted subgraphs obtained via random walks.

Unlike post-hoc explanation methods, GraphOracle integrates the explanation process directly into the GNN training process and models two key dependencies to support class-level self-explainability:
1) \textit{Graph-subgraph dependencies}, by extracting salient subgraphs that act as atomic semantic units \cite{ying2019GNNExplainer,yuan2021on}, and
2) \textit{Subgraph-prediction dependencies}, by learning how these subgraphs contribute to class-level predictions through class-conditional likelihood and an entropy-based regularization objective.

Prior methods often rely on instance-level post-processing \cite{azzolin2023global} or costly prototype search \cite{zhang2022protgnn,seo2023interpretable}, and cannot directly model these dependencies \cite{wang2023gnninterpreter,huang2023global,nian2024globally}. In addition, many of these methods would require solving an NP-hard graph matching problem to align substructures across training instances, making them computationally expensive and limiting their scalability and faithful class-level  explainability.
To this end, GraphOracle introduces two key technical innovations: 
1) \textit{Entropy-regularized subgraph significance}, which promotes sparse, class-discriminative associations that enhance faithfulness of class-level explanations, and
2) \textit{Efficient subgraph discovery,} using random walks instead of Monte Carlo Tree Search, enabling scalable training and subgraph discovery.

\noindent
\textbf{Contributions.} To summarize, our key contributions are:
\begin{enumerate}
    \item \textbf{New problem formulation.} We introduce the task of class-level self-explainability for GNNs, formalized as a graph-subgraph dependency modeling problem in Sec.~\ref{sec:problem_formulation}. This addresses a critical gap: existing methods such as ProtGNN and PGIB neither evaluate whether their identified prototypes generalize across instances of the same class nor verify whether they produce faithful and discriminative class-level explanations.
    
    \item \textbf{Effectiveness.} GraphOracle produces faithful, class-discriminative explanations while maintaining high classification performance. This is supported by an entropy regularization term that encourages sparse, interpretable subgraph-class associations.

    \item  \textbf{Efficiency.} Our training procedure avoids costly graph matching and prototype search via an integrated dependency-based selection approach, enabling scalable training with speed improvements of up to 12.76 times. 
\end{enumerate}

\section{Related Work}
\label{relatedwork}
\subsection{Instance-level Explanability in GNNs}
Instance-level explainers aim to provide instance-specific explanations for GNN predictions. 
Post-hoc approaches learn a separate explainer for a trained GNN \cite{ye2023SAME,huang2024factorized,ji2024stratified,pmlr-v235-bui24b,jia2025is}.
Recent surveys \cite{yuan2023survey,kakkad2023survey} categorize these methods into five types: gradient-based \cite{baldassarre2019explainability,pope2019explainability}, perturbation-based \cite{ying2019GNNExplainer,luo2020parameterized,yuan2021on,schlichtkrull2021interpreting,hu2025gafexplainer}, surrogate-based \cite{vu2020PGM,huang2023GraphLIME},  decomposition-based \cite{schnake2022higher,xiong2023relevant}, and generation-based approaches \cite{shan2021reinforcement,li2023dag,lin2021generative}. These methods typically assess input features' importance for a specific instance or employ interpretable surrogate models to approximate predictions. 
In contrast, self-explainable GNNs \cite{yu2021graph,miao2022interpretable,wu2022discovering,liu2022graph,sui2022causal,deng2024self} integrate explanation into the training process. 
However, most focus on individual-specific explanations, leaving class-level explainability (high-level understanding across classes) largely unexplored.

\subsection{Class-level Explainability in GNNs}
Class-level explainability studies the general behavior of the model across classes.
Few class-level GNN explainers exist, with most operating post-hoc on pre-trained GNNs.
XGNN \cite{yuan2020XGNN} pioneers this field by training a graph generator to produce patterns that maximize the likelihood of the target prediction made by the GNN. \cite{wang2021towards} encodes class-wise knowledge but struggles to map it to human-interpretable substructures.
To address XGNN's reliance on domain knowledge, \cite{wang2023gnninterpreter} proposes a probabilistic generative method without such requirements. \cite{huang2023global} employs random walks for counterfactual reasoning, while \cite{azzolin2023global} clusters instance-specific explanations from PGExplainer \cite{luo2020parameterized} for class-level insights. \cite{nian2024globally} studies the training utility of class-specific explanations.  
In contrast,  \cite{zhang2022protgnn} and \cite{seo2023interpretable} present self-explanation pipelines, training GNNs while learning prototype significance, and \cite{wang2024unveiling} tries to explore interactive patterns across graphs during training. However, these approaches face challenges in producing faithful class-specific explanations efficiently.

\section{Background and Problem Description}
\subsection{Graph Neural Networks}
A graph is formally represented as a tuple $G = \{V, E, A, X\}$, where $V = \{v_1, v_2, ..., v_n\}$ is the set of nodes (or vertices), $E \subseteq V \times V$ denotes the set of edges; and $A \in \mathbb{R}^{n \times n}$ is the adjacency matrix representing the graph's structure, where $a_{ij} = 1$ if an edge $e_{ij} \in E$ connects nodes $i$ and $j$ $\in V$, and $a_{ij} = 0$ otherwise. The node feature matrix $X \in \mathbb{R}^{n \times k}$ contains $k$-dimensional feature vectors for each of the $n$ nodes. 
A GNN model operates on the fundamental principle of iterative neighborhood aggregation to learn the node embeddings for a graph $G$. This process, known as {\it message-passing}, can be formalized as:
\begin{equation}
\label{eq:agg}
    H^l = f_{AGG}(H^{l-1}, A, X),
\end{equation}
where $H^{l}$ denotes the node embeddings in the $l$-th GNN layer, and $f_{AGG}(\cdot)$ is the aggregation function. The process is initialized with $H^0 = X$. 
This message-passing framework forms the basis for various GNN architectures, allowing for flexible and powerful graph representation learning.

\subsection{Class-level Self-explainability in GNNs}
\label{sec:problem_formulation}
Class-level self-explainability in GNNs aims to design a framework with a built-in explanatory design that can intrinsically provide insights into the model's decision-making process across all predictions. This approach addresses the fundamental question: {\it what patterns or graph structures characterize predictions across different classes?}
One solution is to investigate subgraph patterns that serve as fundamental building blocks of graphs, such as functional groups found in molecular graphs. 
Formally, our goal is to develop a GNN self-explainer that maps a GNN model's predictions to a set of explanatory components. This can be expressed as: $\Phi: f_\text{GNN}(\mathcal{G}) \rightarrow \{W, \mathcal{SG}\}$, where $f_\text{GNN}(\cdot)$ is the GNN model, $\mathcal{G} = \{G_1, G_2, ..., G_n\}$ is a set of input graphs, $W$ is a weight matrix, and $\mathcal{SG}$ is a set of salient subgraphs. The explainer $\Phi$ seeks to assess the importance of subgraphs $\mathcal{SG}$ to the predictions made by $f_\text{GNN}$ across the input graphs. Specifically, the weight matrix $W \in \mathbb{R}^{|\mathcal{C}| \times |\mathcal{SG}|}$ corresponds to the set of target classes $\mathcal{C}$. Each element $w_{ij} \in W$ quantifies the dependency of class $C_i \in \mathcal{C}$ on subgraph $SG_j \in \mathcal{SG}$. Higher values of $w_{ij}$ signify stronger relevance of $SG_j$ for predicting class $C_i$.

\section{Methodology}
To handle the critical challenges faced by current approaches on class-level self-explainability in GNNs (as discussed in Sec.~\ref{sec:intro}), we propose a novel self-explainable learning framework that effectively and efficiently discriminates dependencies between critical subgraphs and GNN predictions across classes.
This framework systematically explores two key types of dependencies: 1) graph-subgraph dependencies and 2) subgraph-prediction dependencies, as introduced in Sec.~\ref{sec:intro}.

In this section, we present a comprehensive approach to developing our self-explainable GNN framework, GraphOracle. Our methodology comprises three key components: 
1) \textbf{Subgraph Extraction:} We begin by introducing an efficient and effective method for extracting meaningful subgraphs from the input dataset, which provides a foundation for interpretable explanations.
2) \textbf{GNN-based Embedding via Message-passing Layers:} 
We then elaborate on the design of advanced message-passing GNN layers tailored for graph-level embedding. These layers produce rich, discriminative graph representations.
3) \textbf{Subgraph Dependency Explainer:} Building upon the extracted subgraphs and graph embeddings, we introduce our novel subgraph dependency explainer. 
The overall algorithm is given in Sec.~\ref{sec:algorithm}, followed by computational complexity analysis in Sec.~\ref{sec:complexity}.

\begin{figure*}
    \centering
    \includegraphics[width=.95\linewidth]{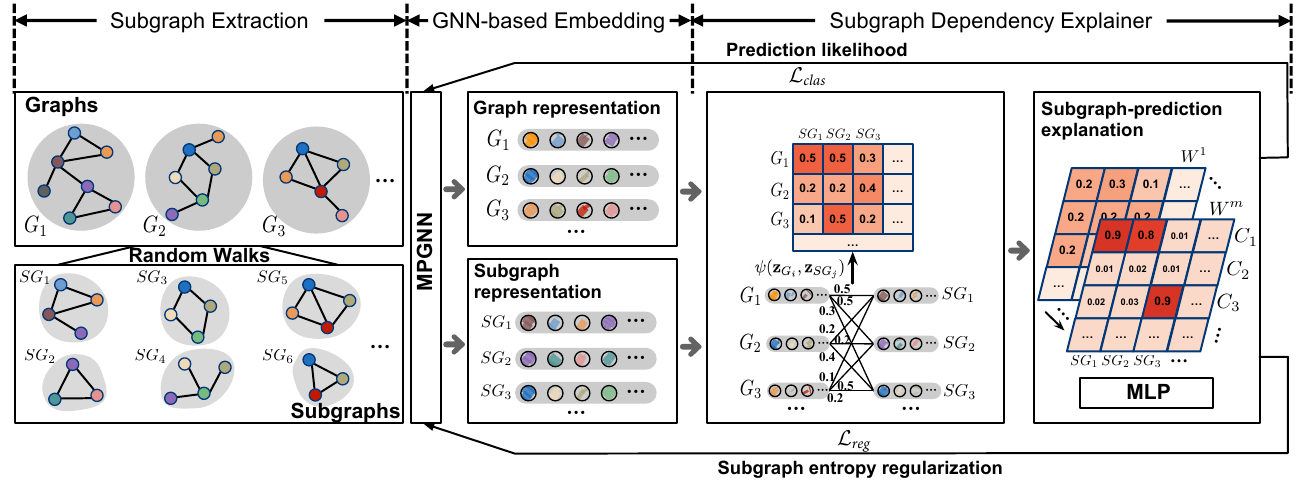}
     \caption{GraphOracle Framework. Subgraph extraction via adaptive random walks, followed by graph and subgraph representation learning. An MLP transforms features to capture graph-subgraph dependencies, with the final layer, $W^m$,  of the weight matrix $W$ capturing class-specific significant subgraph patterns.}
    \label{fig:framework}
\end{figure*}

\subsection{Subgraph Extraction}
\label{sec:subgraph_extraction}
Subgraphs derived from the input graph set play a key role in explaining the GNN model from a global perspective. To extract these subgraphs, we employ a subgraph extractor based on adaptive random walks, which effectively captures informative topological information. 
Our experiments in Sec.~\ref{sec:effectiveness} demonstrate that GraphOracle with random walks outperforms competitors in providing class-level explanations.
Given the unique topological structure of graph data, we adopt an approach similar to \cite{zhao2022from} for our extractor. It samples rooted subgraphs by performing adaptive random walks starting from each node \cite{grover2016node2vec}. For each random walk, we add an induced subgraph of visited nodes to a subgraph set $\mathcal{SG}$, which is then used for training and explaining the GNN. To avoid redundancy, we omit duplicate induced subgraphs that cover the same node set from a graph. 
To mitigate memory overhead and computational complexity, we impose two constraints: 1) the length of each random walk $l_{rw}$ is fixed, and 2) within each batch, random walks are initiated exclusively from a restricted subset of nodes—specifically, the top $K$ nodes ranked in descending order of degree centrality \cite{freeman1978centrality}.

\subsection{GNN-based Embedding}
To achieve high-quality graph representations, we employ GNN-based embeddings that can effectively capture and distinguish the nuances among different graph instances. This can be formalized as:
\begin{equation}
\label{eq:gnn}
    \mathbf{z}_{G_i} = f_\text{GNN}(A_i, X_i),
\end{equation}
where $\mathbf{z}_{G_i} \in \mathbb{R}^d$ is the $d$-dimensional embedding vector for graph $G_i$. The GNN $f_\text{GNN}$ is designed to be at least as powerful as the Weisfeiler-Lehman graph isomorphism test \cite{leman1968reduction}, ensuring that it can distinguish between non-isomorphic graphs and capture structural information effectively. 

To generate discriminative intermediate embeddings for both the original graphs $\mathcal{G}$ and the extracted subgraphs $\mathcal{SG}$, we employ the popular Graph Isomorphism Network (GIN) layers~\cite{xu2018how} for their powerful representational capabilities in capturing graph structure. The embedding process for a graph $G_i$ can be described in two steps: 
1) Node Embedding: We first learn embeddings for all nodes in $G_i$ using the message-passing framework described earlier in Eq.~(\ref{eq:agg}), and  
2) Graph-level Embedding: We then apply a permutation-invariant pooling operation (e.g., sum, mean, or max) to aggregate the node embeddings $H_{(G_i)} \in \mathbb{R}^{|V_i| \times d}$ into the graph-level embedding $\mathbf{z}_{G_i} \in \mathbb{R}^{d}$, where $|V_i|$ signifies the number of nodes in $G_i$, and $d$ denotes the dimension of the last GNN layer. 

\subsection{Subgraph Dependency Explainer}
GraphOracle's self-explainable architecture learns dependencies between target classes and extracted subgraphs through a multi-stage process. 
First, a dependency mapping function transforms graph and subgraph embeddings from a message-passing GNN space to a subgraph-dependency characterized space.
Fully-connected layers then process these dependency-based embeddings to optimize class prediction.
As training proceeds, the model finally captures subgraph-class relationships, providing a class-specific explanation of the GNN's general behavior.

\subsubsection{Subgraph Dependency Learning}
\label{sec:dependency_learning}
Understanding how original graphs depend on extracted subgraphs reveals correlations between class membership and these subgraphs. 
GraphOracle's subgraph dependency learning avoids complex and resource-intensive subgraph matching, enhancing model prediction in a self-explainable manner. We define the subgraph dependency based on GNN-based embeddings: 

\noindent\textbf{Definition} (Subgraph Dependency)\textbf{.} \textit{The subgraph dependency of graph $G_i$ on subgraph ${SG}_j$ is defined as:}
\begin{equation}
\label{kernel}
    f_D(G_i, {SG}_j, f_\text{GNN}) = \psi(\mathbf{z}_{G_i}, \mathbf{z}_{{SG}_j}),
\end{equation}
\textit{where $\mathbf{z}_{G_i}$ and $\mathbf{z}_{{SG}_j}$ are embeddings of $G_i$ and $SG_j$ learned by a message-passing GNN model $f_\text{GNN}$, and $\psi(\cdot)$ measures the information dependency of $G_i$ on ${SG}_j$ in the GNN-based embedding space}.

To implement this definition, we adopt the radial basis function (RBF) kernel \cite{scholkopf2002learning}, which can learn nonlinear decision boundaries benefiting classification performance \cite{kriege2020survey} (analyzed in Appendix~\ref{sec:visualization_kernel}). The dependency of input graph $G_i$ on subgraph $SG_i$ is computed by:
\begin{equation}
\label{eq:rbf}
    \psi(\mathbf{z}_{G_i}, \mathbf{z}_{{SG}_j}) = \exp\left(-\frac{\|\mathbf{z}_{G_i} - \mathbf{z}_{{SG}_j}\|^2}{2\theta^2}\right),
\end{equation}
where $\| \cdot \|^2$ is the squared Euclidean distance and $\theta$ is the RBF kernel's free parameter. 
Graphs $\mathcal{G}$ are then mapped to a new subgraph-dependency characterized space by concatenating their dependencies on all extracted subgraphs $\mathcal{SG}$. The subgraph dependency-based embedding of graph $G_i$ is:
\begin{equation}
\label{eq:concate}
    \mathbf{\tilde{z}}_{G_i} = \text{CONCAT}\left( \{\psi(\mathbf{z}_{G_i}, \mathbf{z}_{{SG}_j}) \ | \ SG_j \in \mathcal{SG} \} \right).
\end{equation}
Next, we model subgraph-prediction dependencies using $m$ fully-connected layers:
\begin{equation} 
\renewcommand{\arraystretch}{1.5}
\label{eq:fc}
    \mathbf{\tilde{z}}_{G_i}^l = 
    \begin{cases}
        W^{l} \cdot \mathbf{\tilde{z}}_{G_i}^{l-1} + b^l, & \text{if } l = m \\[6pt]
        \sigma(W^{l} \cdot \mathbf{\tilde{z}}_{G_i}^{l-1} + b^l) , & \text{otherwise},
    \end{cases}
\end{equation}
where $W^l$ and $b^l$ denote the trainable weights and bias of the $l$-th layer, and $\sigma(\cdot)$ is an activation function. 

We learn the subgraph dependencies by maximizing the likelihood of correct class assignments of the original graphs. This is equivalent to minimizing the cross-entropy loss. For binary classification, this is:
\begin{equation} 
\label{eq:ce}
    \mathcal{L}_{clas} = -\sum_{i=1}^{|\mathcal{G}|} p(\mathbf{\tilde{z}}_{G_i}^m | \mathbf{y}_i) \log p(\mathbf{\tilde{z}}_{G_i}^m | \mathbf{y}_i),
\end{equation}
where $p(\mathbf{\tilde{z}}_{G_i}^m | \mathbf{y}_i)$ denotes the probability that the model predicts the true class $\mathbf{y}_i$ for graph $G_i$.

\subsubsection{Class-specific Explanation}
The training process learns the weight matrix $W^m \in \mathbb{R}^{|\mathcal{C}| \times |\mathcal{SG}|} $ of the final linear layer, where entry $w_{ij}$ quantifies the dependence of class $C_i$ on subgraph $SG_j$ for predictions. 
For each target class $C_i$ $(i= 1, 2, \dots, |\mathcal{C}|)$, we rank the subgraphs $\mathcal{SG}$ by their dependencies: $\mathcal{SG}_{\text{sort}} = \left\{{SG}_{k}\right\}_{k=1}^{|\mathcal{SG}|}$, where $w_k \geq w_{k+1} \in \left\{w_{ij}\right\}_{j=1}^{|\mathcal{SG}|}$. Here, $\left\{w_{ij}\right\}_{j=1}^{|\mathcal{SG}|}$ is the $i$-th row of $W^m$. Subgraphs with higher dependencies are considered high-quality class-specific explanations, elucidating the GNN model's general prediction behavior.

\subsubsection{Regularization}
\label{sec:regularization}
Effective explanations are expected to demonstrate sparse, discriminative significance with respect to GNN predictions across various target classes. That is, subgraphs showing the larger difference in their importance to different target classes serve as high-quality discriminative explanations. To this end, we introduce an entropy regularization, serving the purpose of enhancing optimization and enforcing adherence to the desired property in the class-specific explanations. In particular, the GNN makes predictions for subgraphs $\mathcal{SG}$ characterized by Eqs.~(\ref{eq:rbf})--(\ref{eq:fc}) as it does for original graphs $\mathcal{G}$, and we regularize the entropy of the class membership by minimizing
\begin{equation} 
\label{eq:reg}
    \mathcal{L}_{reg} = -\frac{1}{|\mathcal{C}|} \sum_{j=1}^{|\mathcal{SG}|} \sum_{i=1}^{|\mathcal{C}|} p( \mathbf{\tilde{z}}_{SG_j}^m | \tilde{y}_i) \log p(\mathbf{\tilde{z}}_{SG_j}^m | \tilde{y}_i)
\end{equation}
where $p( \mathbf{\tilde{z}}_{SG_j}^m | \tilde{y}_i)$ denotes the probability that subgraph $SG_j$ is assigned to class $C_i$. 

\subsubsection{Training Objectives} 
We optimize the combined objectives of subgraph dependency learning and the desired properties of focused class-discriminative explanations via:
\begin{equation}
\label{eq:objective}
    \arg\max_{\Theta} \mathcal{L} = \lambda \mathcal{L}_{clas}+ (1-\lambda) \mathcal{L}_{reg},
\end{equation}
where $\lambda$ balances subgraph dependency learning and regularization, and $\Theta$ encompasses all trainable parameters (including $W^m$). This approach bridges classification-oriented training with discriminative subgraph dependency learning.

\subsection{Algorithm of GraphOracle}
\label{sec:algorithm}
The overall algorithm of GraphOracle is summarized in Algorithm~\ref{alg:algorithm}.
GraphOracle iteratively processes both the input graphs and subgraphs to learn embeddings using GNN layers (Lines 5-7). It then transforms these embeddings into subgraph dependency-based representations (Lines 8-10), which are further processed through fully connected layers. The algorithm uses a Softmax function to compute probabilities for the graphs and subgraphs (Line 12), and optimizes the GNN and fully connected layers based on an objective function (Line 13). This process is repeated for a specified number of iterations to achieve a better performance on the validation set, after which the algorithm returns the weight matrix $W$ of last fully-connected layer that represents the subgraph-prediction dependencies.

\begin{algorithm}[!ht]
\caption{The Algorithm of GraphOracle}
\label{alg:algorithm}
\textbf{Input}: training graphs $\mathcal{G}$ with labels $\mathbf{y}$; subgraph set $\mathcal{SG}$; GNN layers $f_\text{GNN}$; fully connected layers $f_\text{FC}$; maximum training iteration $T$; trade-off parameter $\lambda$\\
\textbf{Output}: matrix $W$
\begin{algorithmic}[1] 
\STATE Initialize associated parameters 
\STATE Let $t=1$ 
\STATE Get the set of unique labels $\tilde{y} = \text{unique}(\mathbf{y})$
\WHILE{$t \leq T $}
\FOR{each graph $G_i \in \mathcal{G}$ or subgraph $SG_j \in \mathcal{SG}$}
\STATE Obtain the embedding $\mathbf{z}_{G_i}$ or $\mathbf{z}_{SG_j}$ by Eq. (\ref{eq:gnn})
\ENDFOR
\FOR{each graph $G_i \in \mathcal{G}$ or subgraph $SG_j \in \mathcal{SG}$}
\STATE Obtain the subgraph dependency-based embedding $\mathbf{\tilde{z}}_{G_i}$ or $\mathbf{\tilde{z}}_{SG_i}$ by Eq.~(\ref{eq:concate})
\STATE Obtain the embedding $\mathbf{\tilde{z}}^m_{G_i}$ or $\mathbf{\tilde{z}}^m_{SG_i}$ by Eq.~(\ref{eq:fc})
\ENDFOR
\STATE Obtain $p(\mathbf{\tilde{z}}_{\mathcal{G}}^m)$ and $p(\mathbf{\tilde{z}}_{\mathcal{SG}}^m)$ by the Softmax function
\STATE Update $f_\text{GNN}$ and $f_\text{FC}$ by optimizing Eq.~(\ref{eq:objective})
\ENDWHILE
\STATE \textbf{return} matrix $W = W^m$
\end{algorithmic}
\end{algorithm}

\subsection{Computational Complexity}
\label{sec:complexity}
The major computational cost of GraphOracle stems from the subgraph extraction, message-passing GNN, graph and subgraph dependency measurement, and MLP for explanation. Specifically, the complexity of our random walk-based subgraph extraction is $\mathcal{O}(l_{rw}|V_r|)$, and the message-passing GNN costs approximately $\mathcal{O}(|\mathcal{G}||\bar{V}|kd + |\mathcal{SG}|l_{rw}kd)$ for graphs and subgraphs through each layer, where $|V_r|$ counts nodes selected as roots for subgraph extraction; $|\bar{V}|$ is the average number of nodes in a single graph. 
Calculating the dependency-based embeddings of graphs and subgraphs takes about $\mathcal{O}(\frac{1}{2}|\mathcal{G}||\mathcal{SG}| + \frac{1}{2}|\mathcal{SG}|^2)$ while the cost of the MLP for graph classification and regularization is $\mathcal{O}(m|\mathcal{G}||\mathcal{SG}| + m|\mathcal{SG}|^2)$, since $|\mathcal{SG}|$ is much larger than the number of classes (output layer dimension). 
Consequently, considering the redundancy of rooted subgraphs, i.e., $|\mathcal{SG}| \leq |V_r|$, 
the overall computational cost of $T$ iterations with 2 GNN layers approximates to 
$\mathcal{O}(2|\mathcal{G}||\bar{V}|kdT + 2|V_r|l_{rw}kdT + m|\mathcal{G}||V_r|T + m|V_r|^2T)$.

\section{Experimental Evaluation}
We rigorously evaluate GraphOracle's class-level explanation capabilities across two synthetic and four real-world datasets from bioinformatics and social networks. This diverse selection tests GraphOracle's performance on various graph structures, node features, and classification tasks. 
Our evaluation focuses on effectiveness and efficiency, and includes an ablation study on regularization, and parameter analyses for objective trade-off, subgraph extraction, and fully-connected layers, followed by an analysis for subgraph dependency kernel selection. These analyses aim to answer the following questions: 

\begin{itemize}
    \item \textbf{Q1:} Does GraphOracle demonstrate more effective performance in class-level explanations?
    \item \textbf{Q2:} Is GraphOracle more efficient in approaching class-level explanations?
    \item \textbf{Q3:} Does the regularization introduced to GraphOracle benefit the explanation results regarding faithfulness? 
    \item \textbf{Q4:} How do the parameters impact the explanation results of GraphOracle?
    \item \textbf{Q5:} How does the choice of subgraph dependency kernel affect GraphOracle?
\end{itemize}

As GraphOracle integrates explanation and GNN training phases, unlike most post-hoc class-level explainers, direct quantitative comparisons are challenging. Therefore, we adopt a two-pronged evaluation strategy: 
1) \textit{Qualitative comparison:} We compare GraphOracle's explanations with those from post-hoc and self-explainable methods, focusing on interpretability, coherence, and domain relevance of the generated explanations. 
2) \textit{Quantitative comparison:} We conduct a quantitative comparison with two recent self-explainable GNN models, aiming to evaluate the capability to provide faithful class-specific explanations. 
Our code is available at \mbox{\url{https://github.com/FanzhenLiu/GraphOracle}}.

\subsection{Experimental Setup}
\label{sec:experiment_setup}
\subsubsection{Datasets}
\label{sec:dataset}
We involve both two synthetic datasets and four real-world datasets from bioinformatics and social networks in evaluating the performance of our framework. Their statistics are summarized in Table \ref{tab:datasets}, and more details are given below. 

\begin{table}[h]
    \centering
    \caption{Statistics of the Datasets}
    \resizebox{\linewidth}{!}{
    \begin{tabular}{cccccc}
    \toprule
    \textbf{Dataset} & \textbf{\#Graphs} & \begin{tabular}[c]{@{}c@{}}\textbf{\#Nodes} \\ \textbf{(Avg.)}\end{tabular} & \begin{tabular}[c]{@{}c@{}}\textbf{\#Edges} \\ \textbf{(Avg.)}\end{tabular}           & \textbf{\#Node features} & \textbf{\#Classes} \\ \midrule
    MUTAG            & 188               & 17.93                   & 17.93                             & 7                    & 2                  \\
    Mutagenicity     & 4,337              & 30.32                   & 30.77                             & 14                   & 2                  \\
    BACE             & 1,513              & 34.09                   & 73.72                             & 9                    & 2                  \\
    BA-2Motifs       & 1,000              & 25                      & 25.48                             & 10                   & 2                  \\
    BA-LRP           & 20,000             & 20                      & 21                                & -                    & 2                  \\ 
    HIN              & 1,760              & 11.68                   & 18.03                             & 5                    & 2                  \\ \bottomrule
    \end{tabular}
    }
    \begin{tablenotes} 
            \item Note: One-hot encoding of atom types is applied as node features \\ for MUTAG and Mutagenicity.
        \end{tablenotes} 
    \label{tab:datasets}
\end{table}

\begin{itemize}
    \item \textbf{Synthetic Datasets.} 1) \textit{BA-2Motifs} \cite{yuan2023survey} is composed of Barab\'{a}si-Albert (BA) graphs, where half of them are attached with a ``house"-shaped motif and the remaining half are attached with a cycle motif. 2) \textit{BA-LRP} \cite{schnake2022higher} contains Barabasi-Albert (BA) graphs evenly distributed across two classes, each generated using different growth models. 
    For two classes of graphs in BA-LRP, each class uses one of the following graph growth models to connect a new node $v$ from a current graph $G$ with a probability $p(v)$.
    \begin{equation}
    p(v) = \frac{Degree(v)}{\sum_{ u \in G}Degree(u)},
    \end{equation}
    \begin{equation}
    p(v) = \frac{Degree(v)^{-1}}{\sum_{ u \in G}Degree(u)^{-1}}.
    \end{equation}
    \item \textbf{Real-world Datasets.} 1) \textit{MUTAG} \cite{debnath1991structure} contains molecules labeled with mutagenic or non-mutagenic, where seven types of atoms are represented as nodes and chemical bonds are depicted as edges. 2) Similar to MUTAG, \textit{Mutagenicity} \cite{kazius2005derivation} consists of more mutagenic or non-mutagenic molecules with 14 types of atoms. 
    3) \textit{BACE} \cite{wu2018MoleculeNet} includes human \text{$\beta$}-secretase 1 (BACE-1) inhibitors as molecule graphs of binary classes. The labels indicate their quantitative IC50 and qualitative binding results. 
    4) \textit{HIN} is a social network describing the face-to-face interactions among medical doctors, nurses, administrative staff, and patients \cite{vanhems2013estimating}. We use the data processed in \cite{azzolin2023global}, which includes 1,760 graphs. Each graph represents an ego network with a radius of 3, centered on each doctor and nurse.
\end{itemize}

\subsubsection{Comparison Models}
\label{sec:comparison_model}
To ensure a fair quantitative evaluation of faithfulness, all competitors should employ the same masking strategy. While our goal is to model the graph-subgraph dependencies, existing strategies that focus on masking subgraphs within the original graph \cite{yuan2023survey,seo2023interpretable} are not applicable to operation on graph-subgraph dependency-based features. 
For post-hoc explainers, accurately locating identified class-specific patterns within individual graphs for masking brings another challenge: graph matching--an NP-hard problem. To bridge the gap, we propose a new subgraph dependency-based masking scheme, wherein subgraphs exhibiting higher dependency differences between classes are selectively removed or retained for generating dependency-based embeddings.
Since post-hoc methods and most self-explainable approaches do not explicitly model dependencies between graphs and class-specific subgraphs as maskable features, we compare GraphOracle with two existing prototype-based class-level GNN self-explainers: ProtGNN \cite{zhang2022protgnn} and PGIB \cite{seo2023interpretable}, which can extract learnable weights of critical substructures as dependencies to inform class predictions.
Specifically, our comparison uses prototypes from ProtGNN and PGIB, and subgraphs from GraphOracle, as both provide class-specific explanations. 
For qualitative comparison, we include two prominent post-hoc class-level explainers, XGNN \cite{yuan2020XGNN} and GLGExplianer \cite{azzolin2023global}, alongside ProtGNN and PGIB. To evaluate classification performance, we compare with additional typical GNN-based models built on GCN \cite{kipf2017semi}, GAT \cite{velickovic2018graph}, GraphSAGE \cite{hamilton2017inductive}, and GIN \cite{xu2018how}.

\subsubsection{Evaluation Metrics} 
\label{sec:metrics}
Given the challenge of evaluating explanation results in the absence of ground truths,
we follow \cite{yuan2023survey} and introduce two metrics to assess the explanatory effectiveness of GraphOracle.

\noindent
\textit{Sparsity.} Effective explanations should highlight the crucial input features driving prediction results while excluding irrelevant ones. To measure this, we use sparsity as a key metric, which quantifies the fraction of features deemed important by explanation techniques. Unlike traditional sparsity metrics that consider edges, nodes, or node attributes as features, GraphOracle computes sparsity based on subgraphs:
\begin{equation}
\label{eq:sparsity}
    Sparsity =  1 - \frac{|\mathcal{SG}^*|}{ |\mathcal{SG}|},
\end{equation}
where $|\mathcal{SG}^*|$ counts the important subgraphs among all $|\mathcal{SG}|$ subgraphs used in the dependency measurement.

\noindent
\textit{Fidelity.} High-quality explanations must be faithful to the prediction model. If an explanation accurately identifies the critical features influencing the model, removing these features should significantly alter its predictions.
Fidelity is thus defined as the variance between the original predictions and those obtained by removing or retaining key features \cite{pope2019explainability}. In line with the faithfulness evaluation framework adopted in \cite{yuan2023survey,hu2025gafexplainer}, the $Fidelity_+$ score quantifying the prediction probability drop is computed as:
\begin{equation}
\label{eq:fidelity_plus}
    Fidelity_+ = \frac{1}{|\mathcal{G}|} \sum_{i=1}^{|\mathcal{G}|} \big[p(G_i|y_i) - p(\tilde{G}_i^{\widehat{\mathcal{SG}}}|y_i)\big], 
\end{equation}
where $\widehat{\mathcal{SG}} = \mathcal{SG} \setminus\mathcal{SG}^*$ denotes the subgraph set considered for dependency-based embeddings after removing the critical ones. $\tilde{G}_i^{\widehat{\mathcal{SG}}}$ refers to the modified version of the original test graph $G_i \in \mathcal{G}$, in which the influence of crucial subgraphs $\mathcal{SG}$ (determined by a sparsity level specified in Eq.~(\ref{eq:sparsity})) has been removed by performing a mask on subgraph dependency-based embeddings. The terms $p(G_i|y_i)$ and $p(\tilde{G}_i^{\widehat{\mathcal{SG}}}|y_i)$ denote the prediction probabilities of class $C_i$ (associated with the ground truth label $y_i$) for graph $G_i$ and $\tilde{G}_i^{\widehat{\mathcal{SG}}}$, respectively, when fed into the trained GNN that learns subgraph/prototype dependencies (e.g., ProtGNN, PGIB, and our GraphOracle). A higher $Fidelity_+$ score, indicating greater changes when dependencies from critical subgraphs are ignored, implies a higher degree of faithfulness in class-level explanations.

Analogously, the $Fidelity_-$ score that assesses the degree to which the predictions approximate the original outputs based on explanatory subgraphs, is calculated as:
\begin{equation}
\label{eq:fidelity_minus}
    Fidelity_- = \frac{1}{|\mathcal{G}|} \sum_{i=1}^{|\mathcal{G}|} \big[p(G_i|y_i) - p(\tilde{G}_i^{\mathcal{SG}^*}|y_i)\big], 
\end{equation}
where $\tilde{G}_i^{\mathcal{SG}^*}$ is derived from the original test graph $G_i$, in which only the dependencies from crucial subgraphs $\mathcal{SG}^*$ have been considered by performing a mask on subgraph dependency-based embeddings. The terms $p(G_i|y_i)$ and $p(\tilde{G}_i^{\mathcal{SG}^*}|y_i)$ denote the prediction probabilities of class $C_i$ (associated with the ground truth label $y_i$) for graph $G_i$ and $\tilde{G}_i^{\mathcal{SG}^*}$, respectively, when fed into the trained GNN that learns subgraph/prototype dependencies. A lower $Fidelity_-$ score, reflecting smaller changes when only considering dependencies from critical subgraphs, indicates the more faithful class-level explanations.

Furthermore, to compare the faithfulness from both perspectives above in a comprehensive manner, the gap between $Fidelity_+$ and $Fidelity_-$ is adopted as:
\begin{equation}
\label{eq:fidelity_delta}
    Fidelity_\Delta =  Fidelity_+ - Fidelity_-. 
\end{equation}
Stronger faithfulness of class-level explanations is therefore reflected by a higher $Fidelity_\Delta$ score.

\subsubsection{Implementation Environment and Hyperparameter Settings}
\label{sec:implementation_setup}
We conduct all the experiments on 1 NVIDIA Hopper H100 (SXM5) with 94GB Memory, within a Dell XE9640 cluster system running Linux. The environment is configured with Python 3.12, CUDA 12.2, PyTorch 2.2.2, and PyTorch Geometric (PYG) 2.5.2. 
To ensure fairness in empirical comparisons, all GNN explainers applied to the same dataset are configured with an identical batch size and a uniform data split of 80\% for training, 10\% for validation, and 10\% for testing. The batch size is set to 150 for MUTAG, 800 for BA-2Motifs, and 512 for the remaining datasets. For the post-hoc approaches (XGNN and GLGExplainer), we utilize the authors’ open-source implementations and employ the pre-trained GNNs as targets for explanation when available for a given dataset. In cases where pre-trained models are not provided, we train a GIN model \cite{xu2018how} instead.
Then, other hyperparameter settings specific for class-level GNN explainers are detailed as follows:
\begin{itemize}
    \item \textbf{GraphOracle}: We employ the optimizer Adam \cite{kingma2015adam} for training to achieve satisfactory performance in terms of explainability and classification. To achieving optimal results, we adopt a grid serach exploring the learning rate in \{0.005, 0.01, 0.02, 0.05, 0.1\}, the random walk length of 9 or 10 with parameters $p$ and $q$ set to \{0.5, 1, 2\}, the hidden dimensions of two GIN layers in \{32, 64, 128, 256\} with ReLU activation, $\lambda$ in the range [0.85, 0.99] with a step of 0.01, and top $K$ values in \{1, 2, 5, 10, 20, 50, 100, 200\} for selecting rooted nodes with highest degrees in each batch. $\theta$ for the RBF kernel is fixed at 1. The number of training epochs is set to 20 for BA-LRP and 200 for all other datasets. For down-stream graph classification, we employ two fully-connected layers with ReLU activation for MUTAG and Mutagenicity, with LeakyReLU for BACE, and with Sigmoid for the others. Sum pooling is used to derive graph-level embeddings from node embeddings, with the input dimension of the fully-connected layers equal to the number of extracted subgraphs $|\mathcal{SG}|$. The dropout ratio is 0.5.

    \item \textbf{XGNN}: We follow the default setting in its official open-sourced implementations for explaining pre-trained GCN-based models. The post-hoc explanation model is trained for 200 epochs. The source code provided by the authors can be found on GitHub\footnote{https://github.com/divelab/DIG/tree/main/dig/xgraph/XGNN}.
    
    \item \textbf{GLGExplainer:} Since the authors only provide instance-level explanations for Mutagenicity \cite{azzolin2023global}, we follow their settings and generate instance-level explanations on other datasets using PGExplainer\footnote{\url{https://github.com/divelab/DIG/tree/main/dig/xgraph/PGExplainer}} \cite{luo2020parameterized} for explaining pre-trained 3-GIN-layer models. For these datasets not covered in their work, we set the number of prototypes equal to the number of classes, and adopt the instance-level explanation selection scheme, that is, ``elbow\_method", outlined in their code. Following the default configures provided by the authors, GLGExplainer is trained for 500 epochs on real-world datasets and 2,000 epochs on synthetic datasets. The source code for GLGExplainer can be found on GitHub\footnote{\url{https://github.com/steveazzolin/gnn_logic_global_expl}}.

    \item \textbf{ProtGNN}: To guarantee the prototype projection works for each dataset at least once, the prototype projection starts from the 50th training epoch for all datasets. 
    All other settings follow those claimed in the original paper \cite{zhang2022protgnn}. The source code for ProtGNN can be found on GitHub\footnote{\url{https://github.com/zaixizhang/ProtGNN/tree/main}}.

    \item \textbf{PGIB:} We follow the default setting in PGIB implementations to explain GIN-based models. The source code provided by the authors can be found on GitHub\footnote{\url{https://github.com/sang-woo-seo/PGIB}}.
\end{itemize}

For other GNN-based models, we follow the parameter settings from \cite{xu2018how} for GCN \cite{kipf2017semi}, GAT \cite{velickovic2018graph}, GraphSAGE \cite{hamilton2017inductive}, and GIN on graph classification, and compare the results achieving the best performance on the validation set with the same data split mentioned above.
For GIN architectured with 4 GNN layers (excluding the input layer), we search the optimal classification performance using the hidden dimensions in \{16, 32\} for molecular graphs (MUTAG, Mutagenicity, and BACE) and 64 for social graphs (BA-2Motifs, BA-LRP, and HIN), the batch size in \{32, 128\}, and the dropout ratio in \{0, 0.5\}.
For the other three, with the hidden dimensions set to 64, the batch size of 128, and the dropout ratio of 0.5, we optimize their performance by varying the number of GNN layers (excluding the input layer) from 2 to 4. For GAT, we employ 8 attention heads as default in \cite{velickovic2018graph}. All models are trained for 350 epochs on graph classification tasks, following the default setting in \cite{xu2018how}, and employ the same data split as the GNN explainers.

\subsection{Effectiveness Evaluation (\textbf{Q1})}
\label{sec:effectiveness}
\subsubsection{Visualization Analysis for Class-specific Explanations} 
\label{sec:visualization}
To evaluate GraphOracle's capability to generate human-understandable explanations, we compare the significant subgraphs it identifies with the class-specific subgraphs/prototypes found by XGNN, GLGExplainer, ProtGNN, and PGIB. 
Unlike our approach and other self-explainable methods that incorporate subgraph/prototype learning directly, XGNN and GLGExplainer perform post-hoc explanations for pre-trained massage-passing GNNs, making direct quantitative comparison with identical metrics challenging.
Therefore, following \cite{yuan2020XGNN} and \cite{azzolin2023global}, we offer a qualitative comparison through visual representations.

\begin{figure}[!tb]
\centering
    \includegraphics[width=1\linewidth]{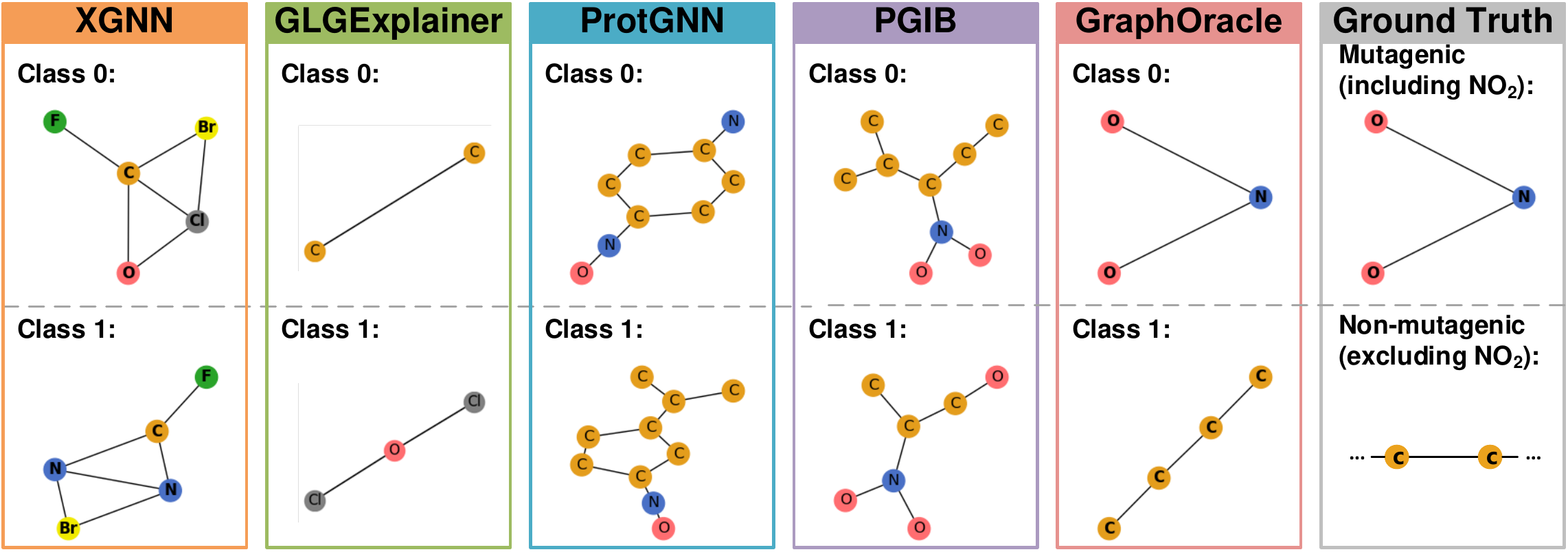}
    \caption{Visualization of the subgraph (or prototype for ProtGNN and PGIB) with the highest probability/weight/dependency for each class in MUTAG.  
    The ground truth explanations include the functional group $NO_2$ for mutagenic molecules and exclude it for non-mutagenic ones.}
    \label{fig:mutag}
\end{figure}

We use the bioinformatics dataset MUTAG and the synthetic social network dataset BA-2Motifs—both known for their reliable ground truth—as case studies for GNN class-specific explanations.
For the proposed GraphOracle, we sort the extracted subgraphs in descending order based on their dependencies for classes, and select the top subgraph for each class in MUTAG and the top four subgraphs for each class in BA-2Motifs.
For XGNN, we set $max\_node$ to 5 for MUTAG, generating one class-specific explanation graph per class, and vary $max\_node$ from 4 to 7 for BA-Motifs, generating four explanation graphs per class.
For GLGExplainer, 
we choose the first generated graph for each class in MUTAG and the first four generated graphs per class in BA-2Motifs, using the inspect function provided by the authors. 
For ProtGNN and PGIB, we select the top prototype for each class in MUTAG and the top four prototypes for each class in BA-2Motifs, based on their associated weights to class predictions. 

\begin{figure}[!tb]
\centering
    \includegraphics[width=1\linewidth]{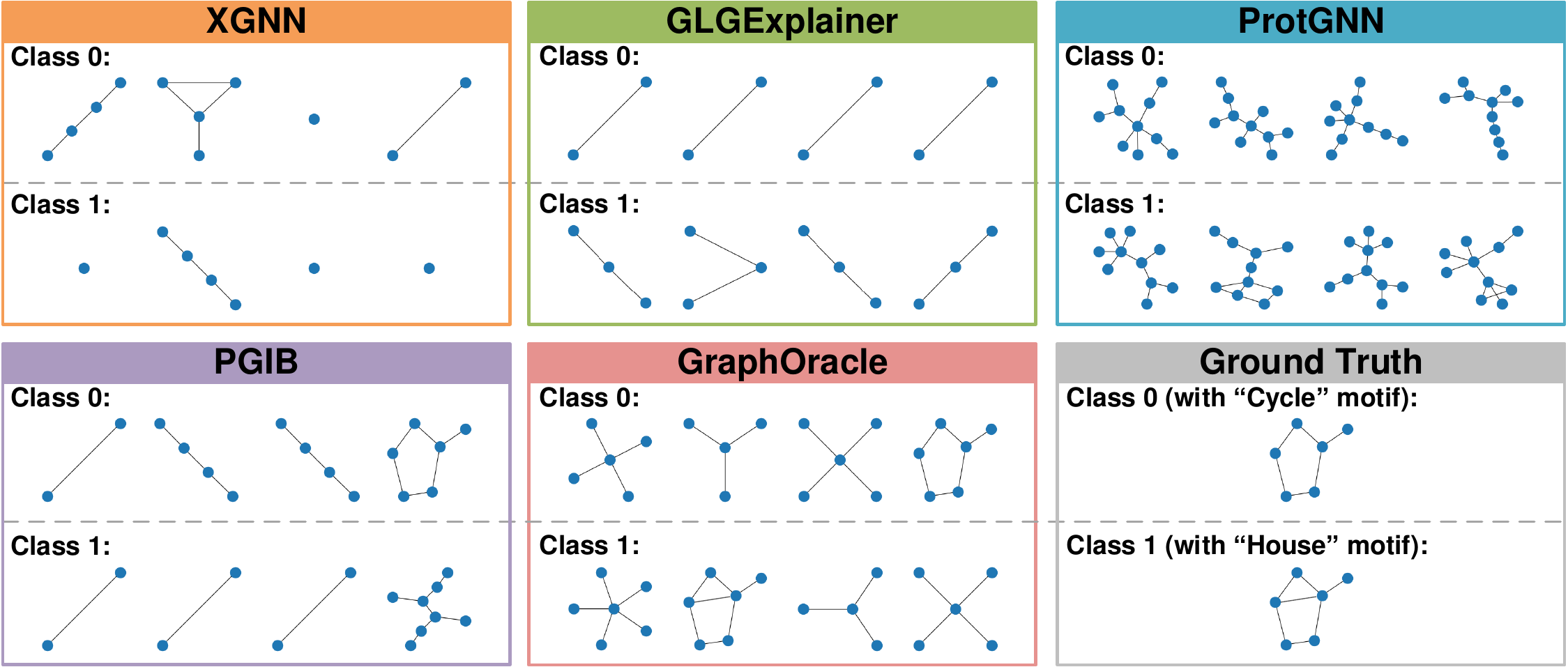}
    \caption{Visualization of the Top 4 subgraphs (or prototypes for ProtGNN and PGIB) with the highest probability/weight/dependency for each class in BA-2Motifs. The ground truth explanations include the ``Cycle'' motif for Class 0 and the ``House'' motif for Class 1.}
 \label{fig:ba2motifs}
\end{figure}

As shown in Fig.~\ref{fig:mutag}, GraphOracle effectively captures the $NO_2$ group as the most significant subgraph for mutagenic molecules and identifies non-mutagenic molecules by highlighting carbon chains excluding $NO_2$. Hence, the class-level explanations provided by GraphOracle align with molecule classification based on the presence of $NO_2$, unlike the other methods which fail to do so.  
In addition, Fig.~\ref{fig:ba2motifs} indicates that ProtGNN includes the ``House" motif together with other uninformative substructures in its second prototype for Class 1, whereas PGIB is able to identify the true ``Cycle" motif for Class 0. In contrast, for both classes, GraphOracle can accurately identify the gold motifs within the top four subgraphs. Both pos-hoc explainers, XGNN and GLGExplianer, fail to produce explanations that align with this fact in BA-2Motifs. 

Consistent with the observations in \cite{seo2023interpretable}, ProtGNN incorrectly incorporates uninformative substructures into graph-level embeddings, missing $NO_2$ as a mutagenic indicator in MUTAG and the ``Cycle" motif reprenstative of Class 0 in BA-2Motifs. PGIB fails to learn sparse weights for discriminative prototypes across different classes, mistakenly associating the $NO_2$ group with the non-mutagenic class and overlooking the ``House" motif as an indicator for Class 1 in BA-2Motifs. 
Compared to its reported success on the Mutagenicity dataset, as observed in its original paper, GLGExplainer's failure on MUTAG and BA-2Motifs can be largely attributed to its reliance on class-specific patterns derived from instance-level explanation graphs.

\subsubsection{Faithfulness of Explanations}
High-quality class-level explanations reveal the key factors driving GNN predictions for different classes. To evaluate the faithfulness of explanations provided by dependency-architectured GNN self-explainers, we train all explainers using the same data split and then mask out or retain the subgraphs/prototypes with the highest class-specific dependencies/weights under various sparsity levels on the test set.

As clarified in Section~\ref{sec:experiment_setup}, existing instance-specific masking strategies \cite{seo2023interpretable,yuan2023survey} are not suitable for evaluating class-specific explanations, and post-hoc explainers face the NP-hard challenge of graph matching. To address these limitations, we propose a feature-based masking strategy that operates on graph-subgraph dependencies, selecting the subgraphs/prototypes with the highest variation in dependencies across classes and masking out or retaining the corresponding features within the graph-subgraph dependency-based embeddings in the test set. The number of selected subgraphs/prototypes is determined by the sparsity level as defined in Eq.~(\ref{eq:sparsity}).

\begin{table}[htbp]
\centering
\caption{Average $Fidelity_\Delta$ in percentage across various sparsity levels over five runs. The best results are in \textbf{bold}.}
\label{tab:fid_delta}
\scalebox{0.83}{
\begin{tabular}{cccccccc}
\toprule
\multirow{2}{*}{\textbf{Dataset}} & \multirow{2}{*}{\textbf{Explainer}} & \multicolumn{6}{c}{\textbf{Sparsity}} \\ \cmidrule{3-8} 
                              &             & 0.50           & 0.55           & 0.60           & 0.65           & 0.70           & 0.75           \\ \midrule
\multirow{3}{*}{MUTAG}        & ProtGNN     & 0.50           & -0.05          & -0.05          & -1.56          & -1.56          & -3.63          \\
                              & PGIB        & -21.87         & -20.92         & -22.27         & -19.92         & -22.26         & -21.18         \\
                              & GraphOracle & \textbf{32.74} & \textbf{32.71} & \textbf{32.56} & \textbf{32.34} & \textbf{32.30} & \textbf{32.20} \\ \midrule
\multirow{3}{*}{Mutagenicity} & ProtGNN     & -0.07          & -0.09          & -0.09          & -0.11          & -0.11          & -0.10          \\
                              & PGIB        & -10.17         & -11.10         & -10.95         & -11.79         & -12.84         & -12.65         \\
                              & GraphOracle & \textbf{20.91} & \textbf{20.91} & \textbf{20.86} & \textbf{20.85} & \textbf{20.86} & \textbf{20.82} \\ \midrule
\multirow{3}{*}{BACE}         & ProtGNN     & 4.28           & -3.03          & -3.03          & -3.17          & -3.17          & -8.23          \\
                              & PGIB        & 3.88           & 1.91           & 0.87           & -0.76          & -1.96          & 0.17           \\
                              & GraphOracle & \textbf{26.21} & \textbf{26.21} & \textbf{26.06} & \textbf{25.77} & \textbf{25.72} & \textbf{25.66} \\ \midrule
\multirow{3}{*}{BA-2Motifs}   & ProtGNN     & -4.60          & -5.50          & -5.50          & -8.90          & -8.90          & -10.51         \\
                              & PGIB        & -1.29          & -1.96          & -1.29          & -3.02          & -2.51          & -2.60          \\
                              & GraphOracle & \textbf{26.47} & \textbf{24.76} & \textbf{23.25} & \textbf{20.69} & \textbf{17.83} & \textbf{12.61} \\ \midrule
\multirow{3}{*}{BA-LRP}       & ProtGNN     & 0.01           & 0.01           & 0.01           & 0.00           & 0.00           & -0.01          \\
                              & PGIB        & -44.03         & -44.31         & -43.94         & -44.07         & -43.92         & -43.94         \\
                              & GraphOracle & \textbf{29.64} & \textbf{28.65} & \textbf{24.27} & \textbf{22.57} & \textbf{20.47} & \textbf{13.27} \\ \midrule
\multirow{3}{*}{HIN}          & ProtGNN     & -0.23          & -0.35          & -0.35          & -0.45          & -0.45          & -0.52          \\
                              & PGIB        & -31.13         & -31.39         & -31.64         & -30.41         & -31.92         & -31.43         \\
                              & GraphOracle & \textbf{26.71} & \textbf{25.51} & \textbf{23.66} & \textbf{20.96} & \textbf{18.15} & \textbf{13.96} \\ \bottomrule
\end{tabular}
}
\end{table}

\begin{table}[htbp]
\centering
\caption{Average $Fidelity_+$ in percentage across various sparsity levels over five runs. The best results are in \textbf{bold}.}
\label{tab:fid_plus}
\scalebox{0.83}{
\begin{tabular}{cccccccc}
\toprule
\multirow{2}{*}{\textbf{Dataset}} & \multirow{2}{*}{\textbf{Explainer}} & \multicolumn{6}{c}{\textbf{Sparsity}} \\ \cmidrule{3-8} 
 &  & 0.50 & 0.55 & 0.60 & 0.65 & 0.70 & 0.75 \\ \midrule
\multirow{3}{*}{MUTAG} & ProtGNN & 3.89 & 3.43 & 3.43 & 2.55 & 2.55 & 1.36 \\
 & PGIB & -0.34 & 0.56 & -0.75 & 1.59 & -0.76 & 0.34 \\
 & GraphOracle & \textbf{32.80} & \textbf{32.77} & \textbf{32.66} & \textbf{32.49} & \textbf{32.45} & \textbf{32.38} \\ \midrule
\multirow{3}{*}{Mutagenicity} & ProtGNN & 0.01 & 0.00 & 0.00 & 0.00 & 0.00 & 0.00 \\
 & PGIB & 4.10 & 3.08 & 3.24 & 2.35 & 1.50 & 1.72 \\
 & GraphOracle & \textbf{20.91} & \textbf{20.91} & \textbf{20.86} & \textbf{20.85} & \textbf{20.86} & \textbf{20.82} \\ \midrule
\multirow{3}{*}{BACE} & ProtGNN & 12.75 & 6.14 & 6.14 & 6.00 & 6.00 & 3.61 \\
 & PGIB & 5.52 & 4.47 & 4.02 & 3.30 & 2.69 & 3.55 \\
 & GraphOracle & \textbf{26.22} & \textbf{26.22} & \textbf{26.09} & \textbf{25.83} & \textbf{25.80} & \textbf{25.74} \\ \midrule
\multirow{3}{*}{BA-2Motifs} & ProtGNN & 7.67 & 6.42 & 6.42 & 3.73 & 3.73 & 2.84 \\
 & PGIB & -0.17 & -0.39 & -0.11 & -0.95 & -0.48 & -0.60 \\
 & GraphOracle & \textbf{26.62} & \textbf{24.78} & \textbf{23.68} & \textbf{22.19} & \textbf{20.92} & \textbf{18.28} \\ \midrule
\multirow{3}{*}{BA-LRP} & ProtGNN & 0.01 & 0.01 & 0.01 & 0.01 & 0.01 & 0.00 \\
 & PGIB & -0.06 & -0.38 & -0.02 & -0.20 & -0.05 & -0.12 \\
 & GraphOracle & \textbf{27.13} & \textbf{26.06} & \textbf{24.51} & \textbf{22.18} & \textbf{19.86} & \textbf{16.84} \\ \midrule
\multirow{3}{*}{HIN} & ProtGNN & 0.17 & 0.11 & 0.11 & 0.06 & 0.06 & 0.02 \\
 & PGIB & 0.40 & 0.20 & -0.05 & 1.18 & -0.32 & 0.18 \\
 & GraphOracle & \textbf{27.13} & \textbf{26.06} & \textbf{24.51} & \textbf{22.18} & \textbf{19.86} & \textbf{16.84} \\ \bottomrule
\end{tabular}%
}
\end{table}

\begin{table}[]
\centering
\caption{Average $Fidelity_-$ in percentage across various sparsity levels over five runs. The best results are in \textbf{bold}.}
\label{tab:fid_minus}
\scalebox{0.83}{
\begin{tabular}{cccccccc}
\toprule
\multirow{2}{*}{\textbf{Dataset}} & \multirow{2}{*}{\textbf{Explainer}} & \multicolumn{6}{c}{\textbf{Sparsity}} \\ \cmidrule{3-8} 
 &  & 0.50 & 0.55 & 0.60 & 0.65 & 0.70 & 0.75 \\ \midrule
\multirow{3}{*}{MUTAG} & ProtGNN & 3.40 & 3.48 & 3.48 & 4.11 & 4.11 & 4.99 \\
 & PGIB & 21.53 & 21.48 & 21.52 & 21.51 & 21.50 & 21.52 \\
 & GraphOracle & \textbf{0.06} & \textbf{0.06} & \textbf{0.10} & \textbf{0.15} & \textbf{0.15} & \textbf{0.18} \\ \midrule
\multirow{3}{*}{Mutagenicity} & ProtGNN & 0.09 & 0.10 & 0.10 & 0.10 & 0.10 & 0.10 \\
 & PGIB & 14.26 & 14.18 & 14.19 & 14.14 & 14.33 & 14.37 \\
 & GraphOracle & \textbf{0.00} & \textbf{0.00} & \textbf{0.00} & \textbf{0.00} & \textbf{0.00} & \textbf{0.01} \\ \midrule
\multirow{3}{*}{BACE} & ProtGNN & 8.48 & 9.17 & 9.17 & 9.17 & 9.17 & 11.84 \\
 & PGIB & 1.64 & 2.56 & 3.15 & 4.06 & 4.65 & 3.38 \\
 & GraphOracle & \textbf{0.01} & \textbf{0.01} & \textbf{0.03} & \textbf{0.06} & \textbf{0.07} & \textbf{0.07} \\ \midrule
\multirow{3}{*}{BA-2Motifs} & ProtGNN & 12.27 & 11.92 & 11.92 & 12.64 & 12.64 & 13.35 \\
 & PGIB & 1.12 & 1.57 & 1.18 & 2.07 & \textbf{2.03} & \textbf{2.00} \\
 & GraphOracle & \textbf{0.14} & \textbf{0.02} & \textbf{0.43} & \textbf{1.50} & 3.10 & 5.67 \\ \midrule
\multirow{3}{*}{BA-LRP} & ProtGNN & 0.00 & 0.01 & \textbf{0.01} & \textbf{0.01} & \textbf{0.01} & \textbf{0.02} \\
 & PGIB & 43.97 & 43.93 & 43.92 & 43.87 & 43.87 & 43.82 \\
 & GraphOracle & \textbf{-0.39} & \textbf{-0.09} & 0.18 & 0.98 & 3.92 & 7.26 \\ \midrule
\multirow{3}{*}{HIN} & ProtGNN & \textbf{0.40} & \textbf{0.46} & \textbf{0.46} & \textbf{0.51} & \textbf{0.51} & \textbf{0.54} \\
 & PGIB & 31.53 & 31.59 & 31.59 & 31.58 & 31.59 & 31.62 \\
 & GraphOracle & 0.42 & 0.54 & 0.85 & 1.22 & 1.71 & 2.88 \\ \bottomrule
\end{tabular}%
}
\end{table}

As presented in Tables \ref{tab:fid_delta} to \ref{tab:fid_minus}, GraphOracle performs much better than the other two self-explainable methods, ProtGNN and PGIB, which can provide weighted prototypes as class-level explanations. GraphOracle consistently achieves the highest $Fidelity_\Delta$ and $Fidelity_+$ scores across all six datasets, along with the lowest $Fidelity_-$ scores on three bioinformatics datasets (MUTAG, Mutagenicity, and BACE) at sparsity levels ranging from 0.5 to 0.75. While ProtGNN attains lower $Fidelity_-$ scores than GraphOracle on BA-LRP and HIN, and PGIB achieves lower $Fidelity_-$ scores at higher sparsity levels on BA-2Motifs, both methods exhibit substantially lower $Fidelity_+$ scores (below 0.2\%), indicating their inability to identify class-discriminative prototypes effectively in their prototype-class dependency modeling.
In particular, PGIB's predictions, except on BACE and BA-2Motifs, are heavily driven by instance-specific subgraph patterns, resulting in substantially lower $Fidelity_\Delta$ scores. This behavior stems from its embedding design, which concatenates embeddings derived from instance-specific subgraphs and class-specific prototypes.
Hence, we can conclude that GraphOracle effectively captures more important subgraphs that function as class-level explanations more faithful to its own predictions than those discovered by ProtGNN and PGIB.  
Especially, as the sparsity level goes up to 0.75, which results in fewer subgraph dependency-based features being masked out or retained for faithfulness evaluation, GraphOracle continues to maintain high $Fidelity_\Delta$ and $Fidelity_+$ scores (above 20\%) and low $Fidelity_-$ scores (below 0.2\%) on the three bioinformatics datasets. This insight implies that the top 25\% of extracted subgraphs, identified by their highest dependencies, can effectively drive the class predictions made by GraphOracle.

\begin{figure}[ht]
\centering
    \includegraphics[width=0.40\textwidth]{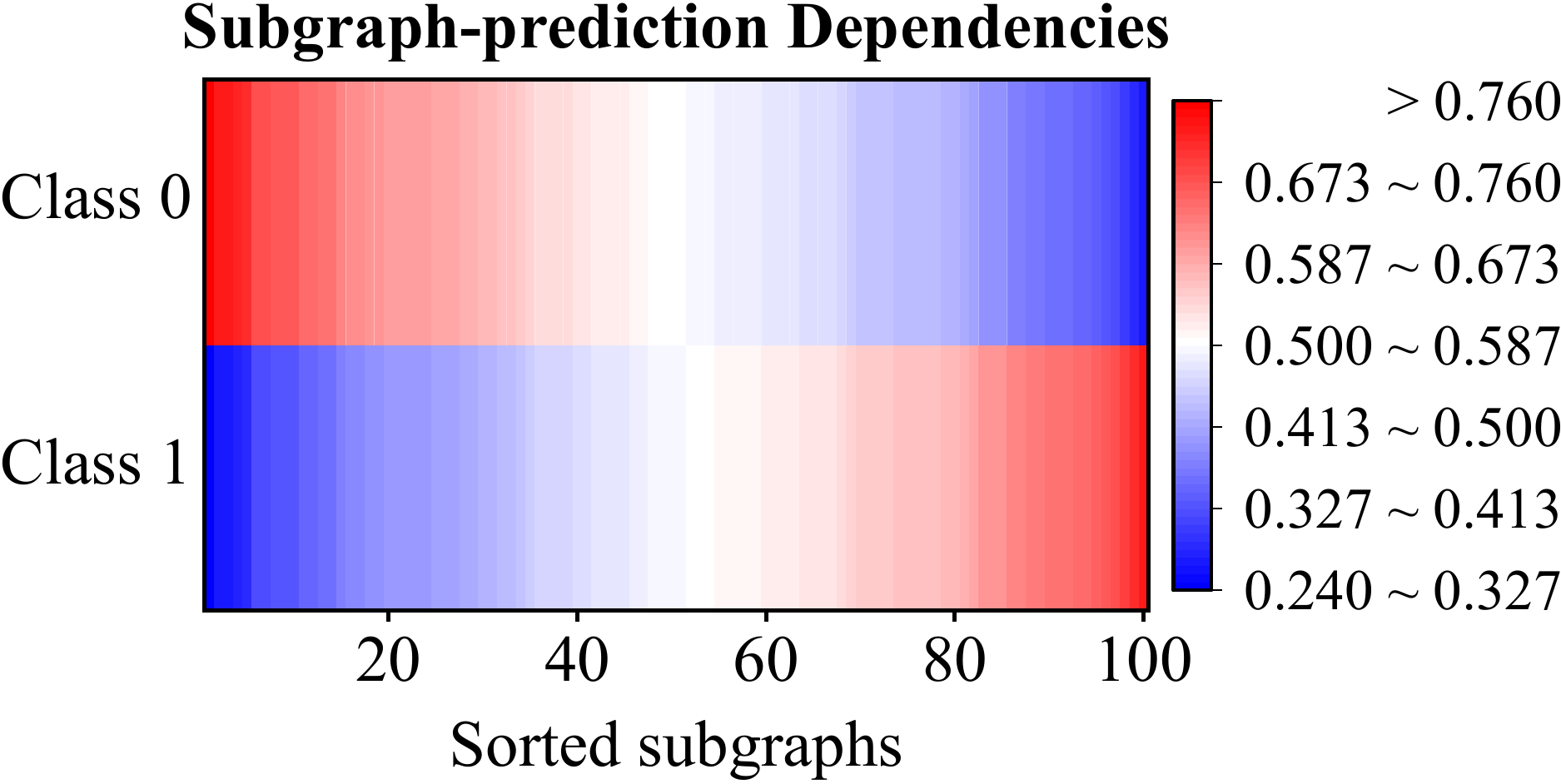}
    \caption{Subgraph-prediction dependencies on MUTAG. The colors depict the degree to which GraphOracle's predictions for each class rely on the extracted subgraphs, and the higher dependencies specific to a class are in dark red. 
    }
    \label{fig:heatmap}
\end{figure}

In addition to the empirical results regarding fidelity and sparsity, we show that GraphOracle is capable of identifying discriminative subgraphs, hereby aligning with the objective of the introduced regularization. 
Taking MUTAG as an example, we visualize the softmax-normalized weight matrix $W^m$ in Fig.~\ref{fig:heatmap}, where 100 pre-extracted subgraphs are ordered by their dependency scores for Class 0. While subgraphs 1 to 10 exhibit the strongest influence on Class 0 predictions (indicated by dark red), subgraphs in the region of 91 to 100 are the most influential for Class 1. This visualization illustrates the sparsity of class-subgraph dependencies—only a small subset of subgraphs strongly contributes to each class. This sparsity, encouraged by the regularization term in Eq.~(\ref{eq:reg}), enables GraphOracle to produce more focused and faithful explanations.

\subsubsection{Classification Performance}
\begin{table}[!tb]
\centering
\caption{Classification accuracy (mean $\pm$ standard deviation) in percentage over five runs, achieved by typical GNN-based models (GCN, GAT, GraphSAGE, and GIN) and self-explainable GNN explainers (ProtGNN, PGIB, and GraphOracle) on training, validation, and test sets. The best results are in \textbf{bold}, the second bests are \ul{underlined}, and the third bests are \colorbox[HTML]{BFBFBF}{shaded}.}
\resizebox{\linewidth}{!}{
\begin{tabular}{ccccc}
\toprule
\textbf{Dataset}               & \textbf{Method} & \textbf{Training}                        & \textbf{Validation}                      & \textbf{Test}                         \\ \midrule
                               & GCN             & 81.87   $\pm$ 1.29                       & 94.44 $\pm$ 0.00                   & 82.00 $\pm$ 4.00                   \\
                               & GAT             & 75.87 $\pm$ 4.98                         & 92.22 $\pm$ 2.72                   & \cellcolor[HTML]{BFBFBF}86.00 $\pm$ 3.74                         \\
                               & GraphSAGE       & \cellcolor[HTML]{BFBFBF}84.40 $\pm$ 1.77                         & 94.44 $\pm$ 0.00                   & 84.00 $\pm$ 3.74 \\
                               & GIN             & \ul{91.87 $\pm$ 1.60}                    & \textbf{100.00 $\pm$ 0.00}               & \ul{88.00 $\pm$ 2.45}                         \\ \cmidrule{2-5}
                               & ProtGNN         & 79.47 $\pm$ 1.71                         & \ul{98.89 $\pm$ 2.22}                   & 78.00 $\pm$ 4.00                         \\
                               & PGIB            & 84.00 $\pm$ 2.89                         & \cellcolor[HTML]{BFBFBF}97.78 $\pm$ 2.72                         & 75.00 $\pm$ 5.48                         \\
\multirow{-7}{*}{MUTAG}        & GraphOracle     & \textbf{92.93 $\pm$ 2.55}                & 96.67 $\pm$ 2.72                   & \textbf{91.00 $\pm$ 2.00}                \\ \midrule
                               & GCN             & 70.67 $\pm$ 1.55                       & 78.20 $\pm$ 1.72                         & 73.52 $\pm$ 1.42                         \\
                               & GAT             & 72.76 $\pm$ 0.87                         & 79.95 $\pm$ 0.82                         & 72.18 $\pm$ 0.52                         \\
                               & GraphSAGE       & 78.27 $\pm$ 2.18                         & 81.48 $\pm$ 1.98                      & 76.14 $\pm$ 1.84  \\
                               & GIN             & \textbf{87.86 $\pm$ 3.49}                & \textbf{86.70 $\pm$ 0.38}                & \textbf{81.93 $\pm$ 0.47}                \\
                               \cmidrule{2-5}
                               & ProtGNN         & \ul{85.78 $\pm$ 2.74}                    & \ul{85.82 $\pm$ 0.72}                   & \ul{80.37 $\pm$ 0.78}                   \\
                               & PGIB            & 78.18 $\pm$ 2.23                         & 80.74 $\pm$ 2.06                         & 74.99 $\pm$ 1.67                         \\
\multirow{-7}{*}{Mutagenicity} & GraphOracle     & \cellcolor[HTML]{BFBFBF}84.38 $\pm$ 1.79 & \cellcolor[HTML]{BFBFBF}81.89 $\pm$ 1.36                          & \cellcolor[HTML]{BFBFBF}76.14 $\pm$ 1.83 \\ \midrule
                               & GCN             & 66.83 $\pm$   2.50                       & 75.10 $\pm$ 1.95                         & 74.74 $\pm$ 5.27                         \\
                               & GAT             & 63.22 $\pm$ 3.09                         & 73.25 $\pm$ 1.85                         & 72.24 $\pm$ 1.34                         \\
                               & GraphSAGE       & 69.09 $\pm$ 1.90                         & 76.95 $\pm$ 1.53                         & 76.97 $\pm$ 3.60                         \\
                               & GIN             & \textbf{91.55 $\pm$ 1.72}                & \textbf{86.75 $\pm$ 0.84}                & \ul{83.42 $\pm$ 2.18}                \\
                               \cmidrule{2-5}
                               & ProtGNN         & \cellcolor[HTML]{BFBFBF}82.41 $\pm$ 1.22 & \ul{83.97 $\pm$ 0.88}                   & \cellcolor[HTML]{BFBFBF}79.61 $\pm$ 2.16 \\
                               & PGIB            & 68.64 $\pm$ 4.14                         & 70.99 $\pm$ 1.70                         & 65.39 $\pm$ 1.79                         \\
\multirow{-7}{*}{BACE}         & GraphOracle     & \ul{86.94 $\pm$ 3.46}                    & \cellcolor[HTML]{BFBFBF}80.00 $\pm$ 1.28 & \textbf{83.42 $\pm$ 0.87}                   \\ \midrule
                               & GCN             & 54.50 $\pm$ 1.58                       & 54.60 $\pm$ 1.20                         & 52.20 $\pm$ 0.40                         \\
                               & GAT             & 54.43 $\pm$ 0.55                         & 54.00 $\pm$ 0.00                         & 52.00 $\pm$ 0.00                         \\
                               & GraphSAGE       & 54.10 $\pm$ 0.41                         & 54.00 $\pm$ 0.00                         & 52.00 $\pm$ 0.00                         \\
                               & GIN             & \textbf{100.00 $\pm$ 0.00}               & \textbf{100.00 $\pm$ 0.00}               & \ul{99.80 $\pm$ 0.40}                   \\
                               \cmidrule{2-5}
                               & ProtGNN         & \cellcolor[HTML]{BFBFBF}99.77 $\pm$ 0.23 & \textbf{100.00 $\pm$ 0.00}               & \cellcolor[HTML]{BFBFBF}99.60 $\pm$ 0.49 \\
                               & PGIB            & 50.70 $\pm$ 0.98                         & 58.00 $\pm$ 3.35                         & 54.00 $\pm$ 4.60                         \\
\multirow{-7}{*}{BA-2Motifs}   & GraphOracle     & \ul{99.92 $\pm$ 0.06}                    & \textbf{100.00 $\pm$ 0.00}               & \textbf{100.00 $\pm$ 0.00}               \\ \midrule
                               & GCN             & 97.67 $\pm$ 0.23                       & 98.50 $\pm$ 0.08                         & 98.57 $\pm$ 0.02                         \\
                               & GAT             & 51.11 $\pm$ 0.19                         & 51.15 $\pm$ 0.00                         & 50.20 $\pm$ 0.00                         \\
                               & GraphSAGE       & 51.01 $\pm$ 0.06                         & 51.15 $\pm$ 0.00                         & 50.20 $\pm$ 0.00                         \\
                               & GIN             & \ul{99.97 $\pm$ 0.02} & \textbf{100.00 $\pm$ 0.00}               & \textbf{100.00 $\pm$ 0.00}               \\
                               \cmidrule{2-5}
                               & ProtGNN         & \cellcolor[HTML]{BFBFBF}99.91 $\pm$ 0.09                    & \cellcolor[HTML]{BFBFBF}99.90 $\pm$ 0.03 & \cellcolor[HTML]{BFBFBF}99.80 $\pm$ 0.18 \\
                               & PGIB            & 79.70 $\pm$ 12.81                        & 80.20 $\pm$ 12.74                        & 80.34 $\pm$ 12.96                        \\
\multirow{-7}{*}{BA-LRP}       & GraphOracle     & \textbf{100.00 $\pm$ 0.00}               & \textbf{100.00 $\pm$ 0.00}               & \textbf{100.00 $\pm$ 0.00}               \\ \midrule
                               & GCN             & 90.09   $\pm$ 1.54                       & 87.16 $\pm$ 0.68                         & \cellcolor[HTML]{BFBFBF}86.25 $\pm$ 1.16                         \\
                               & GAT             & 86.79 $\pm$ 1.05                         & 84.66 $\pm$ 0.00                         & 85.23 $\pm$ 0.80                         \\
                               & GraphSAGE       & 90.97 $\pm$ 2.02                         & \cellcolor[HTML]{BFBFBF}87.50 $\pm$ 1.08 & 85.11 $\pm$ 1.04                         \\
                               & GIN             & \ul{97.20 $\pm$ 3.37}                    & \ul{87.61 $\pm$ 1.88}                   & 86.25 $\pm$ 2.20   \\
                               \cmidrule{2-5}
                               & ProtGNN         & 94.77 $\pm$ 1.26                         & 87.39 $\pm$ 0.43                         & 86.14 $\pm$ 1.42                         \\
                               & PGIB            & \cellcolor[HTML]{BFBFBF}97.16 $\pm$ 1.91 & \textbf{88.64 $\pm$ 1.52}                & \ul{86.36 $\pm$ 1.19}                \\
\multirow{-7}{*}{HIN}          & GraphOracle     & \textbf{98.03 $\pm$ 2.39}                & 86.14 $\pm$ 1.71                         & \textbf{88.18 $\pm$ 2.58}               \\ \bottomrule
\end{tabular}
}
\label{tab:acc}
\end{table}

We evaluate GraphOracle's graph classification accuracy against GNN self-explainers ProtGNN and PGIB, and typical GNN models built on GCN, GAT, GraphSAGE, and GIN, with implementation details provided in Sec.~\ref{sec:implementation_setup}. As shown in Table~\ref{tab:acc}, GraphOracle delivers competitive classification performance, outperforming other methods on four out of six datasets, i.e., MUTAG, BA-2Motifs, BA-LRP, and HIN. 
Although GIN achieves better classification results on Mutagenicity and BACE, it lacks self-explainability, highlighting the future challenge of balancing explainability and performance in GNNs.

\subsection{Efficiency Evaluation (\textbf{Q2})}
\label{sec:efficiency}
\begin{figure}[!tb]
\centering
    \includegraphics[width=1\linewidth]{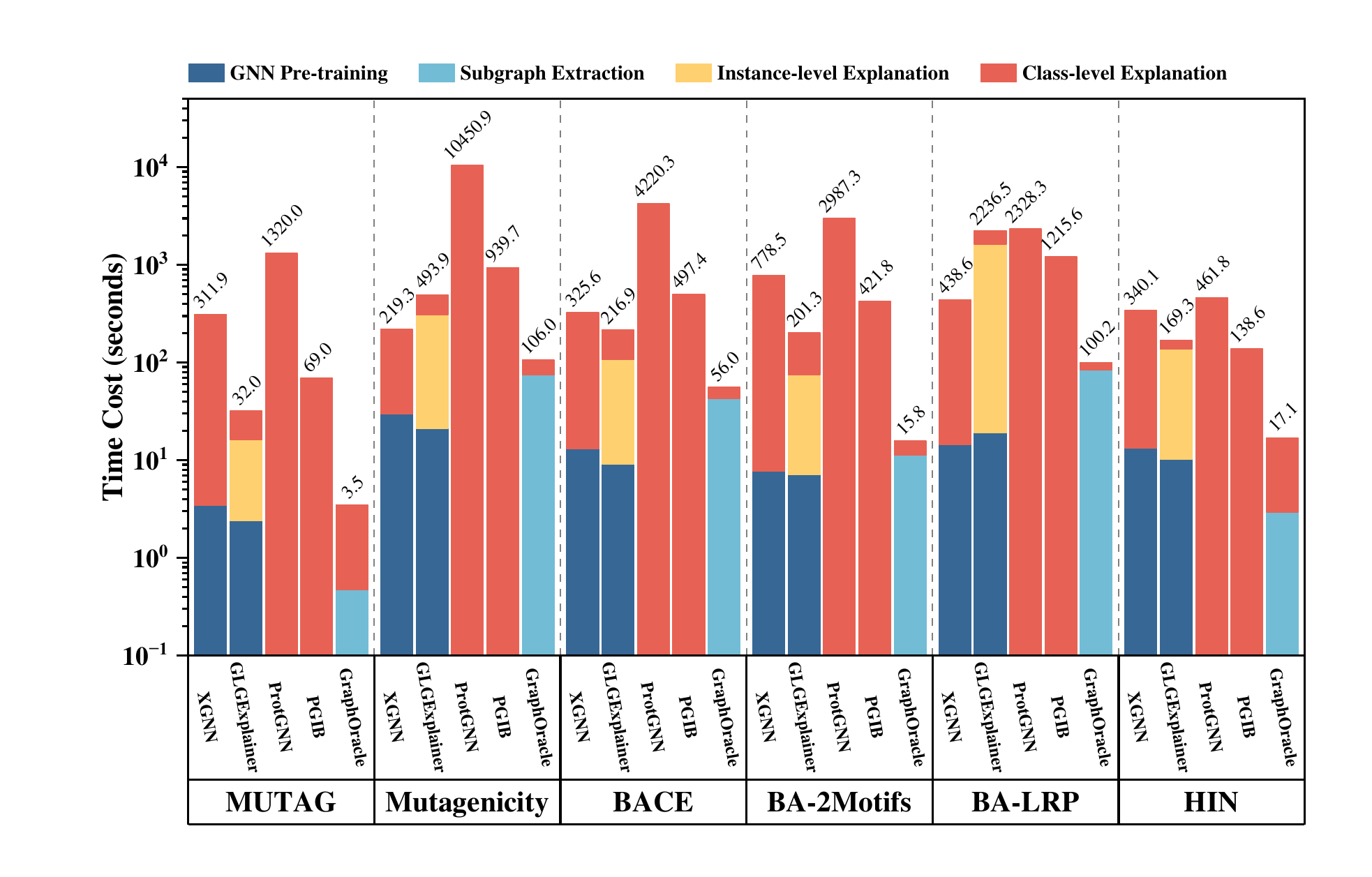}
    \caption{Comparison of average total time cost from pre-training or subgraph extraction to class-level explanation.}
    \label{fig:time_cost}
\end{figure}

We compare the average execution time of different class-level GNN explainers over five runs across multiple datasets, and record the time cost for each phase if applicable. As shown in Fig.~\ref{fig:time_cost}, GraphOracle outperforms all other class-level explanation methods in speed, running at least 2.07 times faster. On the BA-2Motifs dataset, GraphOracle executes up to 12.76 times faster than GLGExplainer, the next fastest option. ProtGNN and PGIB, in contrast, suffer from high computational overhead due to their time-consuming prototype searching during training.

\subsection{Ablation Study on Regularization (\textbf{Q3})}
\begin{figure*}[htbp]
    \centering
    \includegraphics[width=1\textwidth]{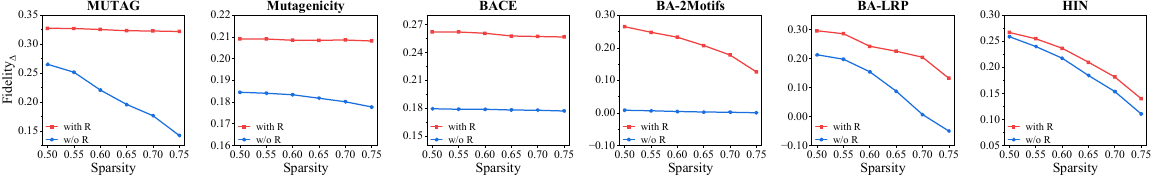}
    \caption{Explanation performance of GraphOracle with (``with R") and without (``w/o R") its entropy regularization, measured by average $Fidelity_\Delta$.}
    \label{fig:ablation}
\end{figure*}

Entropy regularization defined in Eq.~(\ref{eq:reg}) is introduced to identify discriminative subgraphs with sparse correlations across classes. 
We conduct an ablation study to evaluate its effectiveness in enhancing the faithfulness of GraphOracle's self-explainability.
As shown in Fig.~\ref{fig:ablation}, the results across six datasets confirm that the regularization helps GraphOracle capture explanations more faithful to its predictions.

\subsection{Parameter Analyses (\textbf{Q4})}
\label{appx:parameter}
Empirically, we study how the objective trade-off in Eq.~(\ref{eq:objective}), the parameters governing subgraph extraction (Sec.~\ref{sec:subgraph_extraction}), and subgraph dependency measurement (Sec.~\ref{sec:dependency_learning}) impact the class-level self-explanability of GraphOracle, This evaluation assesses the explanations in terms of faithfulness and examines the graph-subgraph dependency-characterized embedding space through visualization.

\subsubsection{$\lambda$ for Objective Trade-off}

\begin{figure*}[ht!]
\centering
    \includegraphics[width=1\textwidth]{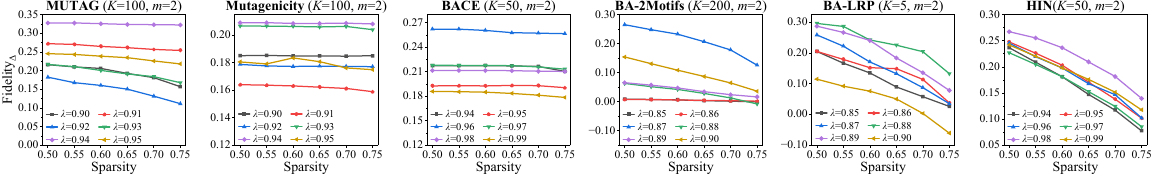}
    \caption{Analysis of training objective trade-off $\lambda$, measured by average $Fidelity_\Delta$.}
    \label{fig:lambda}
\end{figure*}

\begin{figure*}[htbp]
    \centering
    \includegraphics[width=1\textwidth]{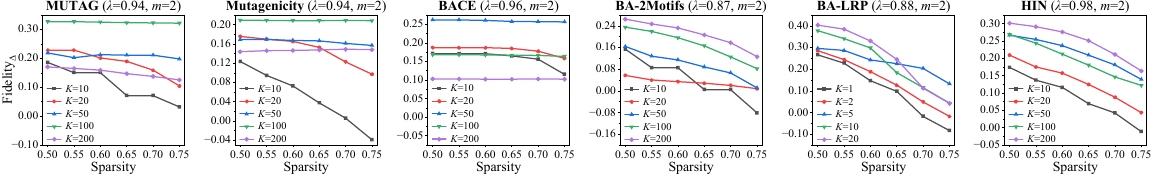}
    \caption{Analysis of top $K$ for subgraph extraction, measured by average $Fidelity_\Delta$.
    }
    \label{fig:topk}
\end{figure*}

\begin{figure*}[htbp]
\centering
    \includegraphics[width=1\textwidth]{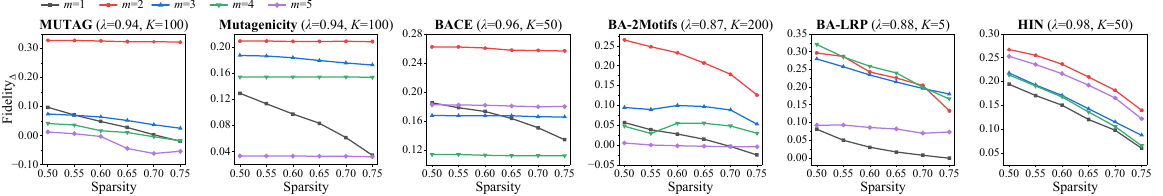}
    \caption{Analysis of the number of fully connected layers $m$ (ranging from 1 to 5), measured by average $Fidelity_\Delta$. 
    }
    \label{fig:layer}
\end{figure*}

Since the subgraph dependency learning and the regularization objectives collaborating in Eq.~(\ref{eq:objective}) are designed to optimize the trade-off between classification performance and the faithfulness of class-level explanations, the choice of $\lambda$ plays a critical role in controlling this balance. To better understand its influence, we perform a sensitivity analysis of $\lambda$ across all six datasets, aiming to identify configurations that yield more faithful explanation results.
As shown in Fig.~\ref{fig:lambda}, at a higher sparsity level, 0.75, for example, GraphOracle achieves more faithful explanations with $\lambda=$ 0.94 on the mutagenic datasets (MUTAG and Mutagenicity), $\lambda=$ 0.96 on BACE, $\lambda=$ 0.87 on BA-Motifs, $\lambda=$ 0.88 on BA-LRP, and $\lambda=$ 0.98 on HIN. Additionally, a smaller value of $\lambda$ tends to encourage sparser subgraph-prediction dependencies, but it may reduce GraphOracle's attention to classification based on graph-subgraph dependencies, ultimately bringing about less faithful explanations due to increased misclassification. This effect is reflected in the fidelity scores, which quantify the shift in model predictions relative to the ground truth. 

\subsubsection{Top $K$ for Subgraph Extraction}
As extracted subgraphs are pivotal in the design of subgraph dependency learning, we investigate the impact of various settings of top $K$ value, which determines the subgraphs in quantity, on class-level explanations delivered by GraphOracle.  
As observed from Fig.~\ref{fig:topk}, GraphOracle is capable of attaining superior class-level explanations in terms of faithfulness measured by the $Fidelity_\Delta$ score, when $K$ is set to 100 for MUTAG and Mutagenicity, 50 for BACE, and 200 for BA-2Motifs and HIN. For BA-LRP, utilizing the top five nodes with highest degrees per batch as roots for subgraph extraction results in more faithful explanation results,
when the sparsity level exceeds 0.65. Additionally, for BA-LRP and HIN, the notable drop in $Fidelity_\Delta$ from a sparsity of 0.70 to 0.75 suggests that subgraphs ranked between the top 25\% and the top 30\% in dependencies are also crucial for model prediction. 
Furthermore, an increase in the number of extracted subgraphs (corresponding to larger $K$ values) does not consistently lead to more faithful explanations provided by GraphOracle, particularly for datasets other than BA-2Motifs.

\subsubsection{Architecture of $m$ Fully-connected Layers}
Given that GraphOracle models the subgraph-prediction dependencies through the learnable weight matrix $W^m$ of the final fully-connected layer in Eq.~(\ref{eq:fc}), we examine the influence of such fully-connected architecture by varying the number of layers $m$ from 1 to 5. The explanation results in Fig.~\ref{fig:layer} suggest that our framework can yield more faithful explainability when employing two fully-connected layer for five datasets (all except BA-LRP) across diverse sparsity levels. For BA-LRP, four layers provide more faithful explanations at sparsity levels up to 0.65, while three layers yield better results at a higher sparsity level of 0.75.

\subsection{Subgraph Dependency Kernel Selection (\textbf{Q5})}
\label{sec:visualization_kernel}

\begin{figure}[!tb]
\centering
    \subfloat[RBF]{
        \includegraphics[width=0.305\columnwidth]{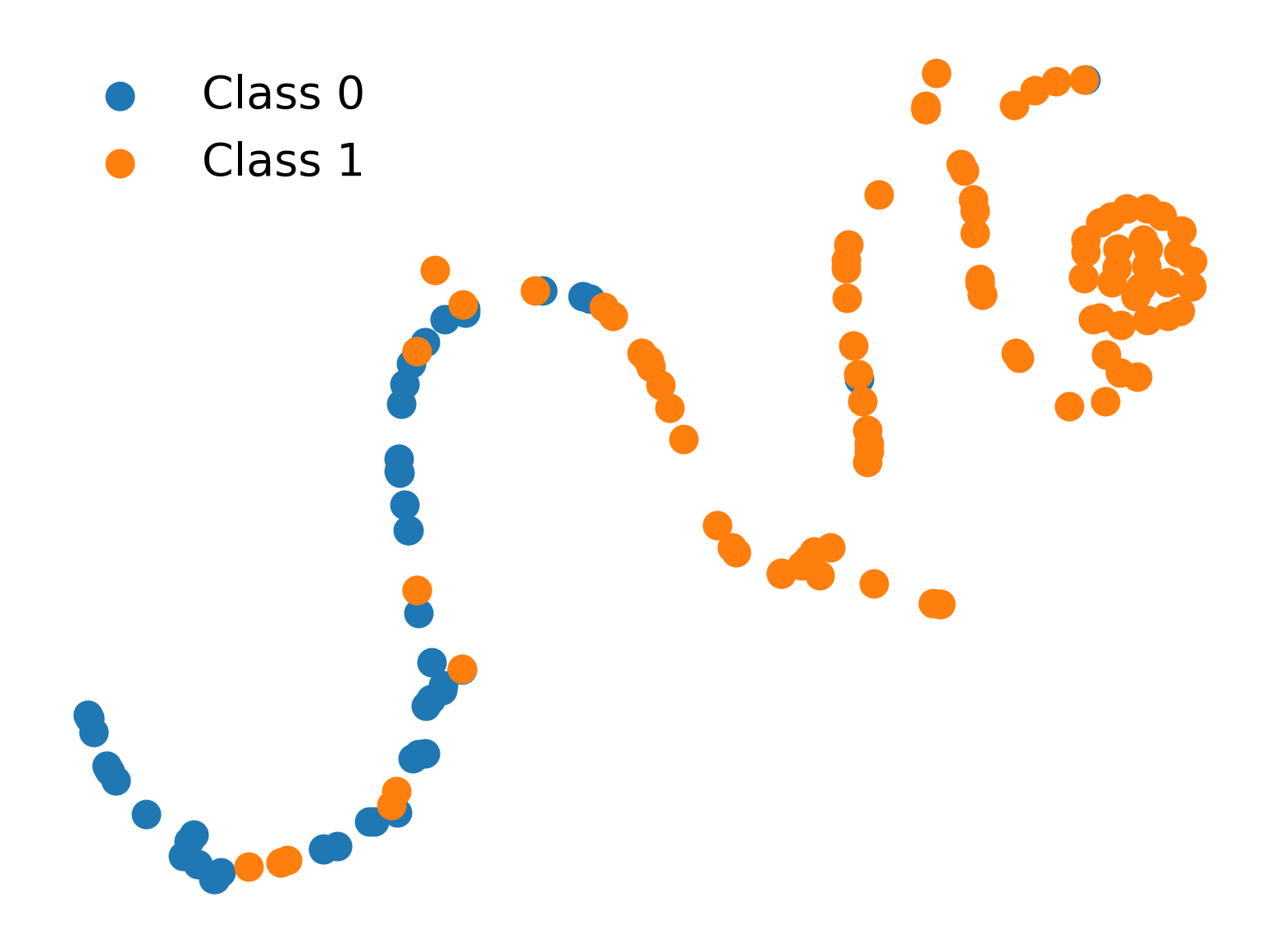}
    }
    \subfloat[Inner Product]{
        \includegraphics[width=0.305\columnwidth]{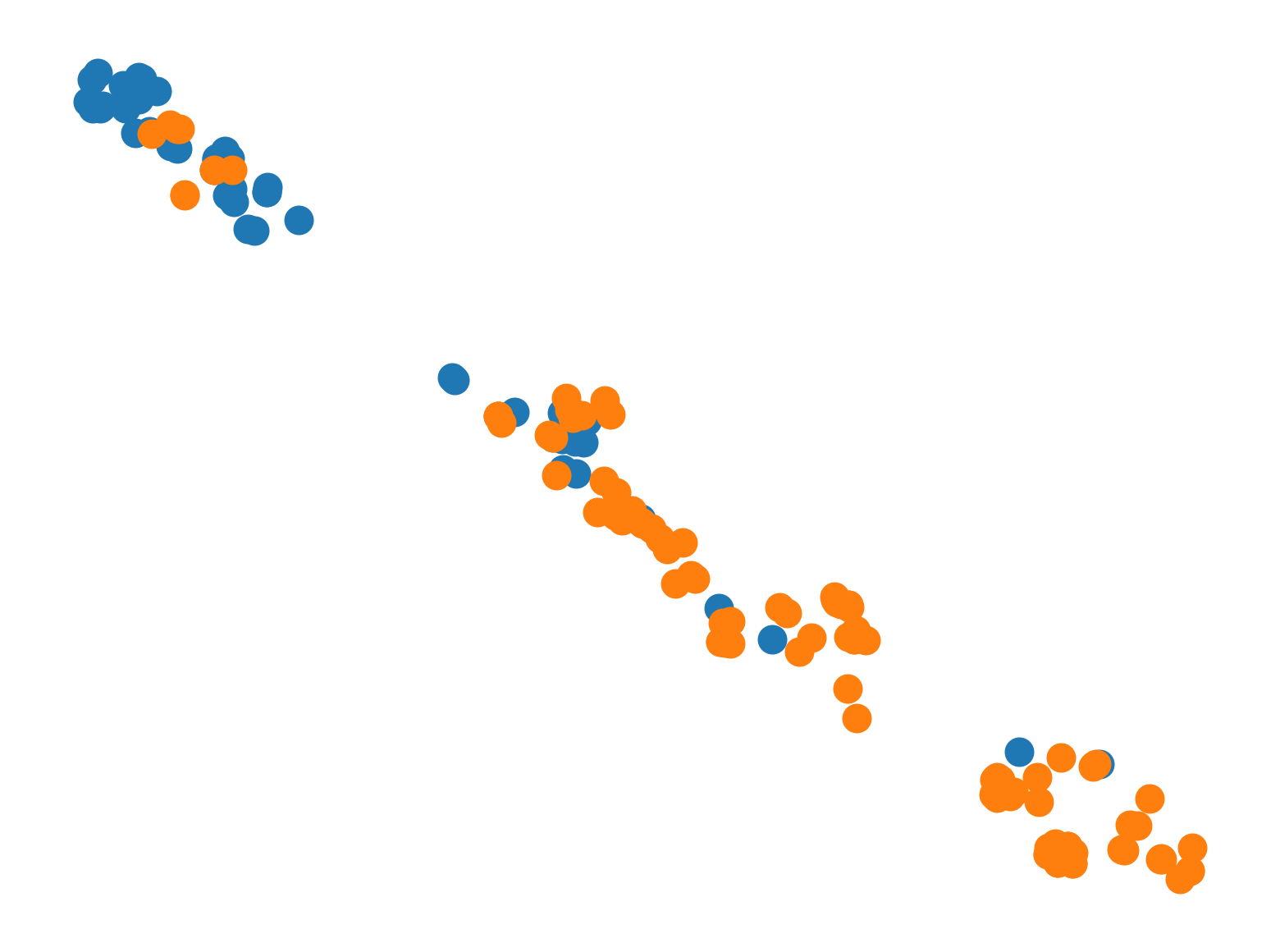}
    }
    \subfloat[Polynomial]{
        \includegraphics[width=0.305\columnwidth]{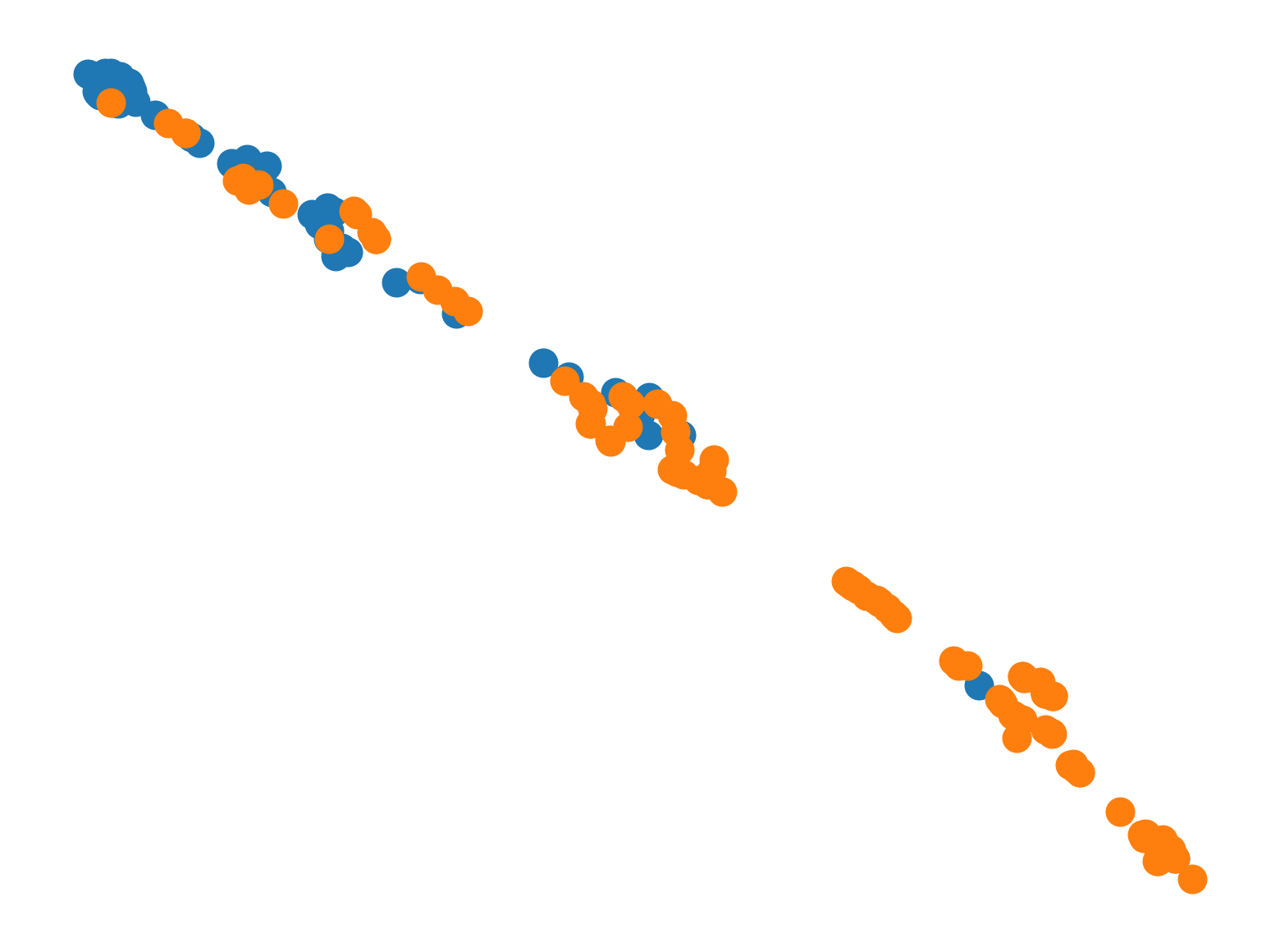}
    }
    \caption{t-SNE visualization of graph-subgraph dependency-based embeddings with diverse kernels on MUTAG.
    }
     \label{fig:visualization_MUTAG_tsne}
\end{figure}

\begin{figure}[!tb]
\centering
    \subfloat[RBF]{
        \includegraphics[width=0.305\columnwidth]{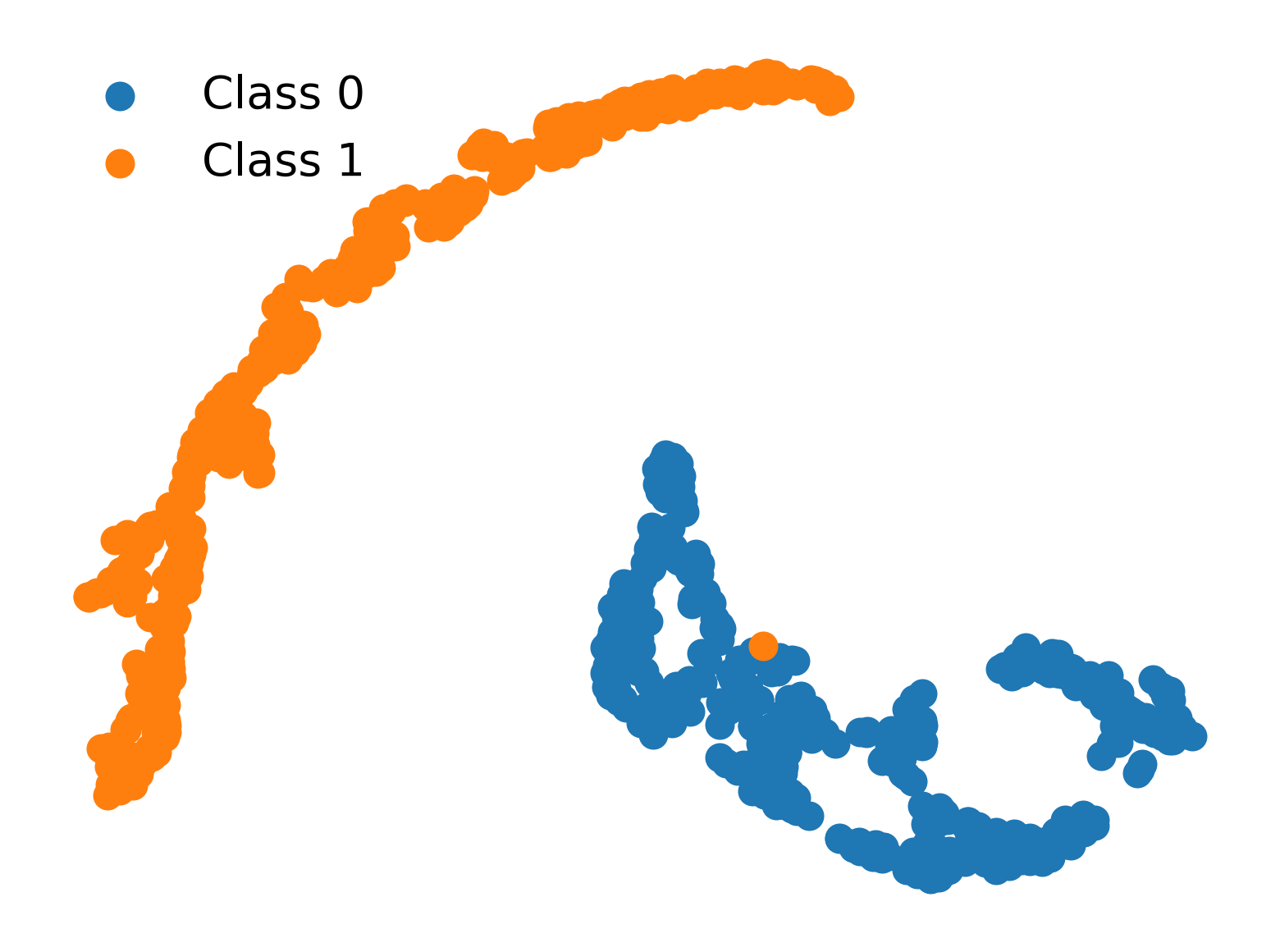}
    }
    \subfloat[Inner Product]{
        \includegraphics[width=0.305\columnwidth]{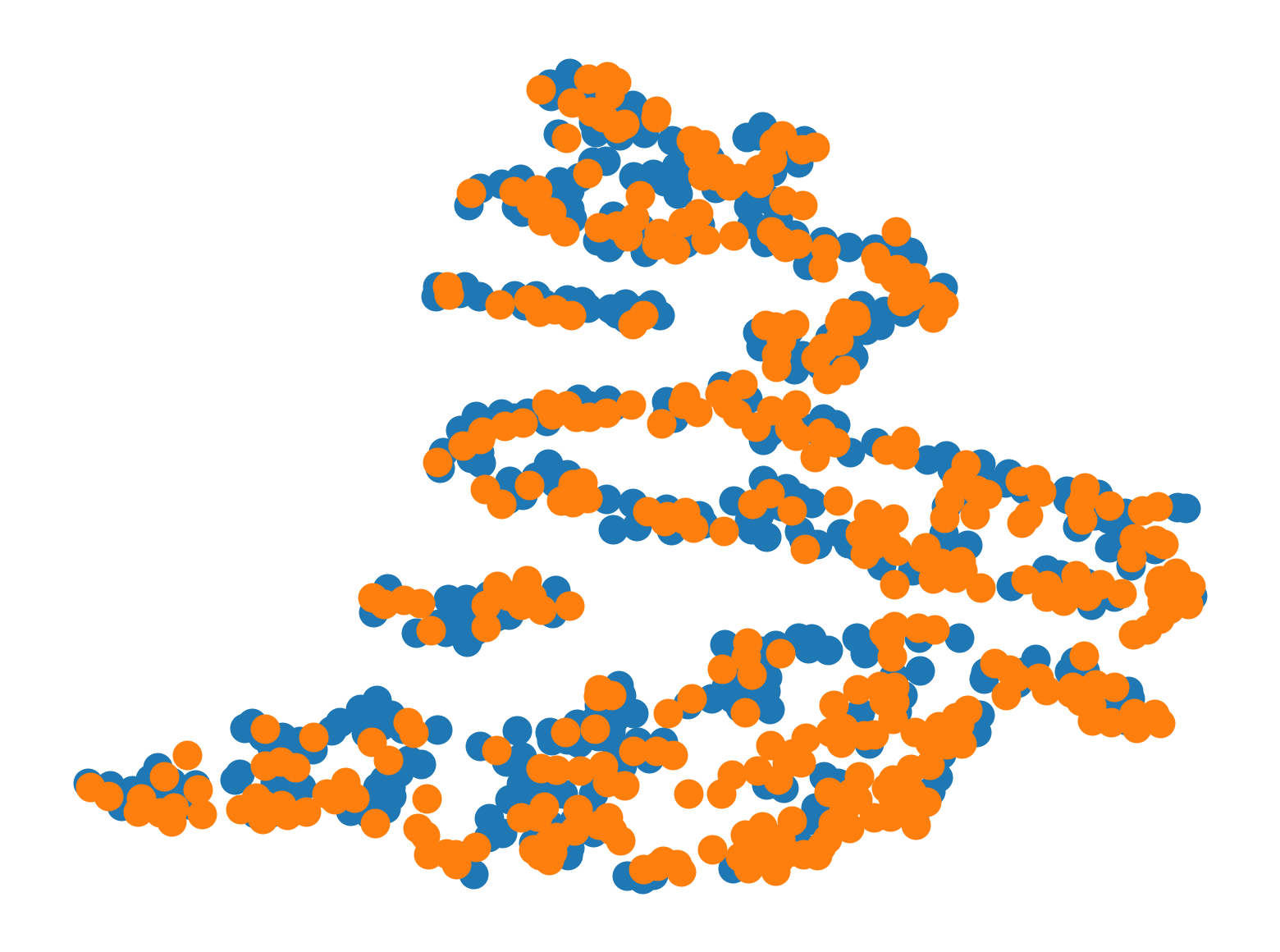}
    }
    \subfloat[Polynomial]{
        \includegraphics[width=0.305\columnwidth]{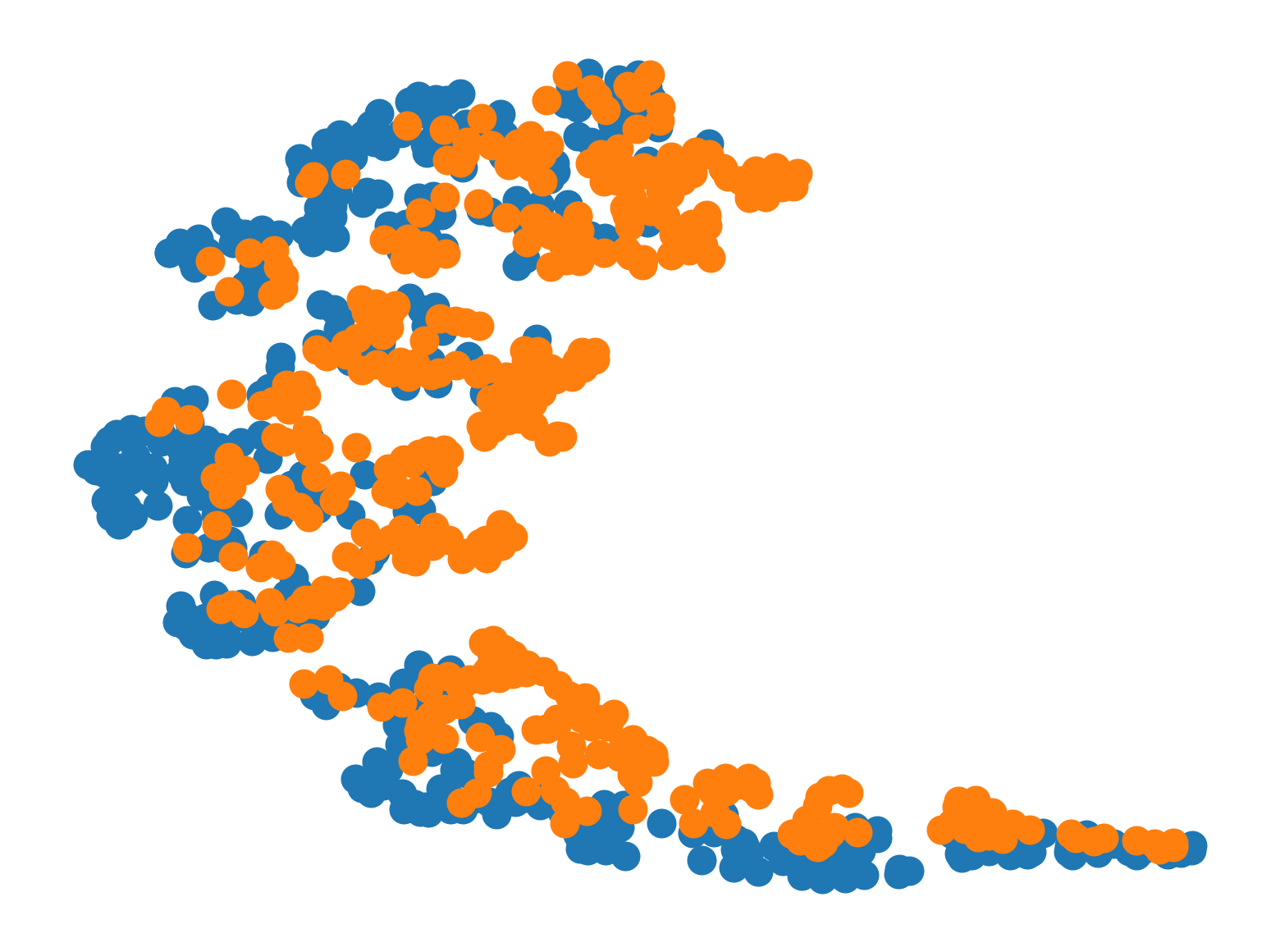}
    }
    \caption{t-SNE visualization of graph-subgraph dependency-based embeddings with diverse kernels on BA-2Motifs. 
    } 
     \label{fig:visualization_BA_2Motifs_tsne}
\end{figure}

To mitigate the impact of the arbitrary arrangement of feature dimensions, it is necessary to employ a measurement that is invariant to permutations when assessing the dependence of a graph on subgraphs.
For this purpose, we incorporate three widely used permutation-invariant kernels \cite{scholkopf2002learning}: the RBF kernel, the Inner Product kernel, and the Polynomial kernel for comparison.
To select the optimal configuration for our GraphOracle, we employ t-SNE visualization \cite{vandermaaten2008tnse1} to explore the subgraph dependency-based embedding space adopting different kernels on a real-world dataset MUTAG and a synthetic dataset BA-2Motifs. As evident from 
Figs.~\ref{fig:visualization_MUTAG_tsne} and \ref{fig:visualization_BA_2Motifs_tsne}, graph instances, demonstrating more pronounced differences between classes in the subgraph dependency-based space measured by the RBF kernel compared to the other two, present significantly enhanced ease for subsequent classification carried out by fully-connected layers. Hence, we equip GraphOracle with the RBF kernel for subgraph dependency learning, pursuing faithful explainability and better performance simultaneously.

\section{Conclusion}
In this paper, we have introduced GraphOracle, a novel self-explainable framework to enhance class-level explainability of GNNs. It uncovers class-specific relevant subgraphs through a graph-subgraph-prediction dependency chain. By integrating GNN training and explanation, GraphOracle overcomes the limitations of post-hoc methods and addresses inefficiencies in current class-level, self-explainable GNN approaches. While it mitigates the trade-off between self-explainability and graph classification task performance, achieving an optimal balance remains an open challenge for future work. 
Furthermore, extending GraphOracle to encompass more complex graph domains, augmenting its capacity for nuanced dependency modeling, and integrating it with emerging GNN architectures offer promising avenues for advancing the frontier of faithful class-level self-explainability.
We believe GraphOracle provides a principled and extensible foundation for future research within the subgraph dependency learning framework.

\bibliographystyle{IEEEtran}
\bibliography{ref}

\end{document}